\documentclass[twoside,11pt]{article}

\usepackage{jmlr2e}
\usepackage{dsfont,rotating,multirow}
\usepackage{amsmath,amssymb}

\usepackage{algorithm,algorithmic}
\newenvironment{varalgorithm}[1]
  {\algorithm\renewcommand{\thealgorithm}{#1}}
  {\endalgorithm}

\newcommand{\ie}{{\em i.e.}, }

\newcommand{\R}{\mathbb{R}}
\renewcommand{\P}{\mathsf{P}}
\newcommand{\E}{\mathsf{E}}
\newcommand{\Var}{\mathsf{Var}}
\newcommand{\err}{\mathsf{Err}(\psi,\psi')}
\newcommand{\1}{\mathds{1}}

\newcommand{\degr}{\textup{deg}}
\newcommand{\vol}{\textup{vol}}
\newcommand{\assoc}{\textup{assoc}}

\newcommand{\Ab}{\mathbf{A}}
\newcommand{\Bb}{\mathbf{B}}

\newcommand{\Ic}{\mathcal{I}}
\newcommand{\Xc}{\mathcal{X}}
\newcommand{\Vc}{\mathcal{V}}

\newcommand{\Ec}{\mathcal{E}}
\newcommand{\Lc}{\mathcal{L}}
\newcommand{\Ac}{\mathcal{A}}
\newcommand{\Dc}{\mathcal{D}}

\newcommand{\Dmin}{\Dc_{\min}}

\newcommand{\Ah}{\widehat{A}}
\newcommand{\Dh}{\widehat{D}}
\newcommand{\Lh}{\widehat{L}}

\jmlrheading{18}{2017}{1-41}{3/16; Revised 3/17}{5/17}{16-100}{Debarghya Ghoshdastidar and Ambedkar Dukkipati}

\ShortHeadings{Uniform Hypergraph Partitioning}{Ghoshdastidar and Dukkipati}
\firstpageno{1}

\begin{document}

\title{Uniform Hypergraph Partitioning: \\Provable Tensor Methods and Sampling Techniques}

\author{\name Debarghya Ghoshdastidar \email debarghya.g@csa.iisc.ernet.in \\
       \name Ambedkar Dukkipati \email ad@csa.iisc.ernet.in \\
       \addr Department of Computer Science \& Automation \\
       Indian Institute of Science\\
       Bangalore - 560012, India}

\editor{Edo Airoldi}

\maketitle

\begin{abstract}
In a series of recent works, we have generalised the consistency results in the stochastic block model literature to the case of uniform and non-uniform hypergraphs. The present paper continues the same line of study, where we focus on partitioning weighted uniform hypergraphs---a problem often encountered in computer vision. This work is motivated by two issues that arise when a hypergraph partitioning approach is used to tackle computer vision problems: 
\\(i)
The uniform hypergraphs constructed for higher-order learning contain all edges, but most have negligible weights. Thus, the adjacency tensor is nearly sparse, and yet, not binary.
\\(ii)
A more serious concern is that standard partitioning algorithms need to compute all edge weights, which is computationally expensive for hypergraphs. This is usually resolved in practice by merging the clustering algorithm with a tensor sampling strategy---an approach that is yet to be analysed rigorously. 

We build on our earlier work on partitioning dense unweighted uniform hypergraphs (Ghoshdastidar and Dukkipati, ICML, 2015), and address the aforementioned issues by proposing provable and efficient partitioning algorithms. Our analysis justifies the empirical success of practical sampling techniques. We also complement our theoretical findings by elaborate empirical comparison of various hypergraph partitioning schemes. 
\end{abstract}


\section{Introduction}

Over several decades, the study of networks or graphs has played a key role in analysing
relational data or pairwise interactions among entities. 
While networks often arise naturally in social or biological contexts,
there are several machine learning algorithms that construct
graphs to capture the similarity among data instances.
A classic example of this approach is
the spectral clustering algorithm of~\citet{Shi_2000_jour_TPAMI} that performs image segmentation 
by partitioning a graph constructed on the image pixels,
where the weighted edges capture the visual similarity of the pixels.
In general, graph partitioning and related problems are quite popular
in unsupervised learning~\citep{Ng_2002_conf_NIPS,Cour_2007_conf_NIPS}, dimensionality reduction~\citep{Zhao_2007_conf_ICML}, 
semi-supervised learning~\citep{Belkin_2004_conf_COLT} as well as transductive inference~\citep{Wang_2008_conf_ICML}.
In spite of the versatility of the graph based approaches, 
these methods are often incapable of handling complex networks
that involve multi-way interactions. 
For instance, consider a market transaction database, where each transaction or purchase corresponds 
to a multi-way connection among commodities involved in the transaction~\citep{Guha_1999_conf_ICDE}.
Such networks do not conform with a traditional graph structure, and need to be modelled as hypergraphs.
Similar multi-way interactions have been considered in the case of molecular interaction networks~\citep{Michoel_2012_jour_PhyRevE}, 
VLSI circuits~\citep{Karypis_2000_jour_VLSI}, tagged social networks~\citep{Ghoshal_2009_jour_PhyRevE}, 
categorical databases~\citep{Gibson_2000_jour_VLDB}, computer vision~\citep{Agarwal_2005_conf_CVPR} among others. 
In this work, we consider the network clustering problem for hypergraphs,
where all the edges are of same cardinality. 
Uniform hypergraph partitioning finds use in computer vision applications
such as subspace clustering~\citep{Agarwal_2005_conf_CVPR,RotaBulo_2013_jour_TPAMI},
geometric grouping~\citep{Govindu_2005_conf_CVPR,Chen_2009_jour_JourCV}
or higher-order matching~\citep{Duchenne_2011_jour_TPAMI}.

Uniform hypergraphs have been in the limelight of theoretical research for
more than a century with the problem of hypergraph colorability surfacing in early $20^{th}$ century~\citep{Bernstein_1908_jour} to recent works
establishing sharp phase transitions in random hypergraphs \citep[see references in][]{Bapst_2015_arxiv_03512}.
However, there has been much less interest in studying practical machine learning problems that deal with uniform hypergraphs. 
For instance, restricting our discussion to the network partitioning,
one can immediately notice a sharp contrast in the theoretical understanding 
of the problem in the context of graph and hypergraphs.
Spectral graph partitioning algorithms have been analysed
from different perspectives since the works of~\citet{McSherry_2001_conf_FOCS} and~\citet{Ng_2002_conf_NIPS}.
The seminal work of~\citet{Rohe_2011_jour_AnnStat}, which studied standard 
spectral clustering under the stochastic block model, drew the attention
of both statisticians and computer scientists, and has led to significant
advancements in understanding of when the partitions can be detected, 
and which algorithms achieve optimal error rates~\citep[see][for the current state of the art]{Abbe_2016_conf_NIPS}.
In contrast, a similar line of study in the context of hypergraphs
is quite recent~\citep{Ghoshdastidar_2014_conf_NIPS,Ghoshdastidar_2015_conf_ICML,Ghoshdastidar_2017_jour_AnnStat,Florescu_2016_conf_COLT}.

\citet{Florescu_2016_conf_COLT} recently solved the 
problem of optimally detecting two equal-sized partitions in a planted 
unweighted hypergraph governed by a single parameter. On the other hand,
the primary focus of our works has been to analyse the consistency of 
hypergraph partitioning approaches used in practice under a more general 
planted model. 
In~\citet{Ghoshdastidar_2014_conf_NIPS}, we presented a generic planted
model for dense uniform hypergraphs, and analysed the tensor decomposition
based clustering algorithm of~\citet{Govindu_2005_conf_CVPR} under this model.
Subsequently, in~\citet{Ghoshdastidar_2015_conf_ICML}, we found that a wide class of so-called ``higher order'' clustering algorithms can be unified by a 
common framework of tensor trace maximisation, which is quite similar in spirit to the associativity
maximisation problem posed in the case of graph partitioning~\citep{Shi_2000_jour_TPAMI}.
We further proposed to solve a relaxation of the problem using a simple,
spectral scheme that was consistent, and achieved better error rates
compared to our previously studied approach.
An extension of the whole setting to the case of sparse non-uniform hypergraphs 
came next on our agenda~\citep{Ghoshdastidar_2017_jour_AnnStat},
and we proved consistency of a spectral approach
for non-uniform hypergraph partitioning.

Like graphs, sparsity turns out to be an important characteristics of 
real-world hypergraphs. While this fact complicates analysis of the algorithms~\citep[see][]{Ghoshdastidar_2017_jour_AnnStat,Florescu_2016_conf_COLT},
it definitely provides significant computational relief.  
For instance, it is easy to realise that for any network clustering scheme,
the computational complexity is at least linear in the number of edges.
Hence, for a $m$-uniform hypergraph on $n$ vertices, any standard approach
should have a $O(n^m)$ runtime unless the hypergraph is sparse.
This is precisely the problem that one encounters in vision applications,
where the network is not given a priori, but one constructs a weighted 
hypergraph using $m$-way similarities among data instances.
Thus, one needs to spend $O(n^m)$ runtime to construct the entire adjacency
tensor only to realise at the end that only few edges have significant
weights, and will aid the partitioning scheme. This scenario motivates 
the study in our present work, where we allow the planted hypergraph to have 
weighted edges, and still be sparse (in the sense that most weights are close to zero).
But, at the same time, the non-zero entries are not known a priori,
and hence, efficient schemes are required to perform the partitioning by 
observing only a small subset of the $O(n^m)$ edge weights.

\subsection{Contributions in this Paper} 
\label{sec_contribution}

We build on our earlier work. To be precise, we study the approach 
presented in~\citet{Ghoshdastidar_2015_conf_ICML}, which solves a relaxation
of the tensor trace maximisation (TTM) problem that lies at the heart of
a variety of higher order learning methods.
On the other hand, the model under consideration is that of sparse
planted uniform hypergraph similar to the one studied in~\citet{Ghoshdastidar_2017_jour_AnnStat}.
However, unlike previous works, we do not restrict the edge weights to be binary, but arbitrary random variables lying in the interval $[0,1]$.
So, the sparsity parameter in our model reduces the mean edge weights,
leading to a large amount of edges with negligibly small weights,
and hence, creating computational challenges of identifying significant edges. 
The planted model is formally described in Section~\ref{sec_model},
while our spectral approach is briefly recapped in Section~\ref{sec_ttm}.
It might come as a surprise to many that this work does not make use
the wide range of tensor decomposition techniques that have now become 
standard tools in machine learning. In Section~\ref{sec_model},
we discuss in detail how our model violates the common structural
assumptions used in the tensor literature.

Our first contribution is presented in Section~\ref{sec_ttm_consistency},
where we analyse the basic TTM approach under the above mentioned planted model for weighted $m$-uniform hypergraphs.
We note that in~\citet{Ghoshdastidar_2015_conf_ICML}, we had studied
 the problem only in the dense unweighted case, whereas similarity
hypergraphs encountered in subspace clustering etc. are weighted and typically
have large number of insignificant edges (sparse).
Furthermore, we recall that spectral partitioning methods, for graphs or hypergraphs, typically require a final step of distance based clustering.
While $k$-means is the practical choice at this stage, theoretical studies
even in the block model literature often use
alternative schemes that are easy to analyse.
Adhering to our goal of studying practical methods, our analysis utilises guarantees of $k$-means algorithm~\citep{Ostrovsky_2012_jour_JACM}
instead of resorting to standard assumptions~\citep{Lei_2015_jour_AnnStat} or other schemes~\citep{Gao_2015_arxiv_03772}.
In this general setting, Theorem~\ref{thm_ttm_consistency} presents the
error rate of the TTM approach under mild restrictions on sparsity.
Furthermore, recent results of~\citet{Florescu_2016_conf_COLT} for the special case of bi-partitioning suggests that our analysis is nearly optimal.
We also show that the performance of this method is similar to the normalised hypergraph cut approach studied in~\citet{Ghoshdastidar_2017_jour_AnnStat},
and is superior than tensor decomposition based partitioning of~\citet{Ghoshdastidar_2014_conf_NIPS}.

The second and key contribution in this paper is the analysis of
a sampled variant of the TTM approach given in Section~\ref{sec_Tetris}. 
As noted above, any basic partitioning scheme would have a $\Omega(n^m)$ runtime
merely due to the construction of the entire adjacency tensor.
We consider a scenario where only $N$ edges are randomly sampled (with replacement) according to some pre-specified distribution.
Theorem~\ref{thm_ttm_consistency_sampled} provides a lower bound on the
sample size $N$ so that the sampled variant achieves desired error rate.
The proof of this result borrows ideas from matrix sampling
techniques~\citep{Drineas_2006_jour_SICOMP}, but mainly relies on a trick of 
rephrasing the problem such that matrix Bernstein inequality can be applied.
The analysis provides quite striking conclusions.
For instance, under a simplified setting, if hypergraph is dense and consists of a constant number of
partitions, then it is sufficient to observe only $\Omega(n(\ln n)^2)$
uniformly sampled edges.
For sparse hypergraphs, uniform sampling cannot improve upon the $\Omega(n^m)$ runtime, but a certain choice of sampling distribution works with only
$N=\Omega(n(\ln n)^2)$.
To the best of our knowledge, this is the first work that analyses graph / hypergraph partitioning with sampling, and such sampling rates 
have not been previously observed in any related tensor problem. 
Typically, most methods need to observe about $\Omega(n^{m/2})$ entries of the tensor in order to estimate its decomposition~\citep[see, for instance,][]{Bhojanapalli_2015_arxiv_05023,Jain_2014_conf_NIPS}, whereas we find that much less observations are required for the purpose of clustering. 
Our analysis also justifies the popularity of the iterative sampling schemes~\citep{Chen_2009_jour_JourCV,Jain_2013_conf_ICCV} in the higher order clustering literature. 

Our final contribution is purely algorithmic. 
We present an iteratively sampled variant of the TTM algorithm,
and conduct an extensive numerical comparison of various methods 
in the context of both hypergraph partitioning and subspace clustering.  
Section~\ref{sec_experiment} presents a wide variety of empirical studies
that (i) validate our theoretical findings regarding relative merits
of TTM over previously studied algorithms,
(ii) compare spectral methods to other hypergraph partitioning algorithms,
including popular hMETIS tool~\citep{Karypis_2000_jour_VLSI},
and (iii) weigh the merits of hypergraph partitioning in subspace clustering
applications, including benchmark problems.
Such empirical studies, though often seen in the subspace clustering works,
was long overdue in the higher order learning literature.
We also hope that the implementations will help 
standardising subsequent studies in this direction.\footnote{Codes are available at: \url{http://sml.csa.iisc.ernet.in/SML/code/Feb16TensorTraceMax.zip} \\and also on the personal webpage of first author.} 

We also note here that to achieve clarity of presentation, the sections
only contain the outline of proofs of the main results. The proofs of the intermediate lemmas and corollaries are given in the appendix that follows after the concluding section (Section~\ref{sec_conclusion}).

\subsection{Notations}
We conclude this section by stating the standard terminology and
notations that we follow in the rest of the paper.
We denote tensors in bold faces ($\Ab,\Bb$ etc.), matrices in capitals ($A,Z$ etc.),
while vectors and scalars will be understood from the context.
We use Trace$(\cdot)$ to write the sum of diagonal entries of a matrix or tensor, and $\Vert\cdot\Vert_2$ 	for Euclidean norm for vector and the spectral norm for matrix.
For a matrix, say $A$, $A_{i\cdot}$ (or $A_{\cdot i}$) represents its $i^{th}$ row (column), $\Vert A\Vert_F$ denotes its Frobenius norm, and
$\lambda_i(A)$ is the $i^{th}$ largest eigenvalue or singular value of $A$ (depending on context).
We also use the standard $O(\cdot)$, $o(\cdot)$ and $\Omega(\cdot)$ notations,
where, unless specified otherwise, the corresponding quantities are viewed as function of $n$. In addition, $\1\{\cdot\}$ is the indicator function, and $\ln(\cdot)$ is natural logarithm.

Moreover, the results in this paper consider two sources of randomness---the random model for hypergraph, and random sampling of edges (or tensor entries).
We make this distinction in the notation for expectation, variance and probability
by specifying the underlying measure. 
For instance, $\E_ H[\cdot]$ is expectation with respect to distribution of the planted model, and
$\E_ {S|H}[\cdot]$ is the expectation over sampling distribution conditioned on a given random hypergraph. Similar subscripts have been used for probability, $\P(\cdot)$, and variance, $\Var(\cdot)$. 
Note that for a matrix $A$, $\E_H[A]$ refers to its entry-wise expectation.

\section{Formal Description of the Problems}
\label{sec_model}
We consider the following random model. 
Let $\Vc = \{1,2,\ldots n\}$ be a set of $n$ nodes, and 
$\psi:\{1,2,\ldots, n\} \to\{1,2,\ldots, k\}$  be a (hidden) partition of the
nodes into $k$ classes. 
For a node $i$, we denote its class by $\psi_i$. 
For a fixed integer $m\geq 2$,
let $\alpha_{n}\in[0,1]$, and $\Bb \in [0,1]^{k \times k \times \ldots \times k}$
be a symmetric $k$-dimensional tensor of order $m$.
Let $\Ec$ be the collection of all subsets of $\Vc$ of size $m$.
A random weighted $m$-uniform hypergraph $(\Vc,\Ec, w)$ is generated
through the random function $w:\Ec\to[0,1]$ such that
\begin{align}
 \E[w(\{i_1,i_2,\ldots,i_m\})] = \alpha_{n}\Bb_{\psi_{i_1}\psi_{i_2}\ldots\psi_{i_m}}
 \label{eq_model}
\end{align}
for all $e=\{i_1,i_2,\ldots, i_m\}\in\Ec$, and the 
collection of random variables $(w(e))_{e\in\Ec}$
are mutually independent.
For convenience, we henceforth write $w_e$ instead of $w(e)$.
The above model extends the planted partition model for graphs~\citep{McSherry_2001_conf_FOCS},
and is a weighted variant of the planted uniform hypergraph model 
studied in our earlier works.
In particular, if $\alpha_{n}=1$ and $w_e$ are independent Bernoulli
random variables satisfying~\eqref{eq_model}, then one retrieves the model of~\citet{Ghoshdastidar_2014_conf_NIPS}.
We present our results for weighted hypergraphs due to 
their extensive use in computer vision~\citep[e.g.,][]{Agarwal_2005_conf_CVPR}.
Note that the edge distributions are governed by $\Bb$ and $\psi$,
and hence, depend only on the class membership.
In addition, $\alpha_{n}$ accounts for sparsity of the hypergraph.
Under this setting, the objective of a hypergraph partitioning algorithm is to estimate $\psi$
from a given random instance of the $m$-uniform hypergraph $(\Vc,\Ec,w)$.
Throughout this paper, we are interested in bounding the error incurred by 
a partitioning algorithm, defined as a multi-class 0-1 loss
\begin{equation}
 \err = \min_\sigma \sum_{i=1}^n \1\{\psi_i\neq\sigma(\psi_i')\}\;,
 \label{eq_error}
\end{equation}
where $\psi$ and $\psi'$ denote the true and the estimated partitions, respectively, and $\sigma$ is any permutation on $\{1,2,\ldots, k\}$.
Note that we also allow number of classes $k$ to grow with $n$.

The first part of the paper builds on the tensor trace maximisation
method of~\citet{Ghoshdastidar_2015_conf_ICML}, and
we prove statistical consistency of TTM under the above sparse planted partition model.
To be precise, we show that under certain conditions on $\alpha_n$ and $\Bb$,
this method achieves $\err=o(n)$.
In the block model terminology~\citep{Mossel_2013_arxiv}, this statement implies that
the algorithm is {weakly consistent}. Furthermore, 
if the hypergraph is dense $(\alpha_{n} = 1)$,  then
we show that TTM can exactly recover the partitions, \ie $\err = o(1)$,
and hence, exhibits strong consistency properties.
From the recent work of \citet{Florescu_2016_conf_COLT}, which studies the special case of bi-partitioning, one can see that our 
restrictions on $\alpha_n$ are nearly optimal (upto a difference of $(\ln n)^2$) in the case of bi-partitioning
as one cannot detect partitions for sparser hypergraphs.

Next we study partitioning algorithms that compute weights of only $N$ out of 
$\binom{n}{m}$ edges. For the theoretical analysis, we assume that there is
a known probability mass function $(p_e)_{e\in\Ec}$, and $N$ edges are 
sampled with replacement from this distribution.
We are interested in finding the minimum $N$ that guarantees weak
consistency of the sampled TTM approach.
We focus on two sampling distributions: (i) uniform sampling, and (ii) weighted sampling where $p_e \propto w_e$.
Surprisingly, we see that if $\alpha_n=1$ (dense case), only 
$N = \Omega(n k^{2m-1}(\ln n)^2)$ edges are sufficient for either sampling strategies. This leads to a drastic improvement in runtime, particularly
if $k=O(1)$, and even in general, since typically $k$ grows much slower than $n$.
However, if $\alpha_n$ decays rapidly with $n$, then more samples
are needed for the case of uniform sampling, whereas the alternative
strategy still works for $N = \Omega(n k^{2m-1}(\ln n)^2)$.
A comparison with tensor literature is not very meaningful, but known methods
of the latter field typically need to observe $\Omega(n^{m/2})$ tensor
entries~\citep{Bhojanapalli_2015_arxiv_05023}. 
In practice, however, computing the weighted sampling distribution
requires a single pass over the adjacency tensor, which still takes $O(n^m)$ time.
But, we argue that practical iterative schemes essentially approximate
this distribution without observing the entire tensor.

\subsection{A Look at Alternative (or Possible) Approaches}
\label{sec_review}

This may be a good time to reflect  on the history of both hypergraphs and tensors
with the focus of understanding the theoretical or practical tools that
either fields provide for the planted uniform hypergraph problem.
Before proceeding further, it will be helpful to take a look at the adjacency tensor of the random (planted) hypergraph.

Let $\Ab \in [0,1]^{n \times n \times \ldots \times n}$ be the symmetric 
adjacency tensor of order $m$. Let $Z\in\{0,1\}^{n\times k}$ be the assignment matrix of the latent partition $\psi$, \ie $Z_{ij} = \1\{\psi_i=j\}$. Then we have
\begin{equation}
\Ab_{i_1 i_2 \ldots i_m} = \left\{
\begin{array}{ll}
 \sum\limits_{j_1,..,j_m=1}^{k} \alpha_n\Bb_{j_1j_2\ldots j_m} Z_{i_1j_1}\ldots Z_{i_m j_m} + \mathbf{E}_{i_1i_2\ldots i_m}
& \text{for distinct } i_1,\ldots,i_m \\
0 & \text{otherwise,} 
\end{array}\right.
\label{eq_A_hosvd_form}
\end{equation}
where $\mathbf{E}$ is a symmetric random tensor with zero mean entries.
For ease of understanding, one may ignore the $O(n^{m-1})$ entries of $\Ab$ with repeated indices to write
\begin{equation}
\Ab \approx \alpha_n(\Bb \times_1 Z \times_2 Z \times_3 \ldots \times_m Z) +\text{noise term},
\end{equation}
where $\times_l$ denotes the mode-$l$ product  between a tensor and a matrix~\citep{Lathauwer_2000_jour_SIAM}.\footnote{ 
 Consider a matrix $B\in\R^{p \times n_l}$ and a $m^{th}$-order
 tensor $\Ab\in\R^{n_1\times n_2\times \ldots \times n_m}$.
 The mode-$l$ product of $\Ab$ and $B$ is a $m^{th}$-order tensor, represented as 
 $\Ab \times_l B \in\R^{n_1 \times \ldots n_{l-1} \times p \times n_{l+1} \times \ldots \times n_m}$,
 whose elements are 
 \begin{center}
  $(\Ab \times_l B)_{i_1 .. i_{l-1} j i_{l+1} ..  i_m} =
  \sum_{i_l}  \Ab_{i_1 .. i_{l-1} i_l i_{l+1} ..  i_m} B_{j i_l}$.
 \end{center}
} 
The basic problem is to detect $Z$ from a given $\Ab$.

The above form is quite similar to the representation of a tensor
in terms of its higher order singular value decomposition or HOSVD~\citep{Lathauwer_2000_jour_SIAM}. In fact, it shows that the random adjacency tensor has a multilinear rank approximately $k\ll n$, and clearly suggests that the partitioning problem should be viewed as a tensor decomposition problem. 
This hint was quickly picked up by~\citet{Govindu_2005_conf_CVPR}, who proposed a spectral approach for higher order clustering based on HOSVD.
Long after this work, tensor methods have gained significant popularity 
in machine learning in recent years.
However, only few works~\citep{Bhaskara_2014_conf_STOC,Anandkumar_2015_conf_COLT} consider decomposition
of tensor into asymmetric rank-one terms, while most of the 
machine learning literature~\citep{Anandkumar_2014_jour_JMLR,Ma_2016_conf_FOCS}
consider decomposition into symmetric rank-one terms.
To be more precise,~\eqref{eq_A_hosvd_form} suggests that
\begin{displaymath}
\Ab \approx \sum_{j_1,\ldots ,j_m=1}^k \alpha_n\Bb_{j_1j_2\ldots j_m} Z_{\cdot j_1} \otimes Z_{\cdot j_2} \otimes \ldots \otimes Z_{\cdot j_m} + \text{noise},
\end{displaymath}
where $\otimes$ is the tensor outer product. Clearly the $k^m$ rank-one terms are asymmetric. It is well known that such a tensor can be represented by a symmetric outer product decomposition only in an algebraically closed field~\citep{Comon_2008_jour_SIAMJMatrixAnal}, \ie one can write as sum of symmetric rank-one terms, but the vectors in the decomposition are not guaranteed to be real, and hence, will be of little use.
We note that though the works of~\citet{Bhaskara_2014_conf_STOC} and~\citet{Anandkumar_2015_conf_COLT} are applicable, their incoherence assumption is clearly violated in the present context where the
same vectors appear in all $m$ modes, and with multiplicity greater than one. 

Under simpler settings such as the one described later 
in Section~\ref{sec_ttm_consistency_example}, one can express $\Ab$ 
as a sum of $(k+1)$ symmetric terms of the form
\begin{equation}
\Ab \approx \sum_{j=1}^k \alpha_n p Z_{\cdot j}^{\otimes m} + \alpha_n q v^{\otimes m} + \text{noise},
\label{eq_A_cp_form}
\end{equation}
where $v=\sum_j Z_{\cdot j}$ and $p, q$ are parameters defining $\Bb$. 
Such tensors, which have a finite symmetric CP-rank (Candecomp/Parafac), have been extensively studied in machine learning. For a single rank-one term, optimal detection rates are known under Gaussian noise~\citep{Richard_2014_conf_NIPS}.
Though there is no distributional assumption on the noise term in our case,~\eqref{eq_model} does imply that the variance of the noise is smaller
than the signal, \ie specialised to this setting, our model lies in the detection region.
Unfortunately, single rank-one term occurs for $k=1$, which does not correspond to any meaningful hypergraph problem, and hence, such results are of little use in our case. 
It may seem that the case of $k\geq2$ can still be tackled using tensor
power iteration based approaches~\citep{Anandkumar_2014_jour_JMLR,Anandkumar_2015_conf_COLT}, 
but the necessary incoherence criterion is violated even here since 
$v$ has a significant overlap with each $Z_{\cdot j}$.
To summarise, it suffices to say that existing guarantees in the tensor literature are not directly applicable for solving~\eqref{eq_A_cp_form},
but it does not rule out the possibility that a careful analysis of power iterations~\citep{Anandkumar_2014_jour_JMLR} or alternative approaches~\citep{Ma_2016_conf_FOCS} may lead to alternative partitioning techniques.
That being said, one should note that~\eqref{eq_A_cp_form} is merely a 
special case of~\eqref{eq_A_hosvd_form}, where HOSVD still appears
to be the natural answer.
 
Interestingly, uniform hypergraphs predate the tensor literature, and one may refer to~\citet{Berge_1984_book_Elsevier}
for early development. Even hypergraph partitioning came into practice~\citep{Schweikert_1979_conf_DesignAutomation} before tensor decompositions gained popularity.
However, initial approaches to hypergraph partitioning in VLSI~\citep{Karypis_2000_jour_VLSI} and database \citep{Gibson_2000_jour_VLDB}
communities relied on clever combination of heuristics with no known performance guarantees.
Subsequent works in computer vision~\citep{Agarwal_2005_conf_CVPR} and machine learning~\citep{Zhou_2007_conf_NIPS} proposed spectral solutions for the problem.
Such approaches are more amenable for a theoretical analysis. While the analysis in~\citet{Ghoshdastidar_2017_jour_AnnStat} and~\citet{Florescu_2016_conf_COLT} are somewhat based on the hypergraph cut approach of~\citet{Zhou_2007_conf_NIPS}, 
the algorithm studied in this paper is closely related to work of~\citet{Agarwal_2005_conf_CVPR}.
The key idea of such spectral schemes is to reduce the hypergraph into 
a graph and then apply spectral clustering. Quite surprisingly, we show in this paper
that both schemes perform better than HOSVD based partitioning both theoretically (Remark~\ref{rem_spectral_comp}) and numerically (Section~\ref{sec_spec_comp}). This is counter-intuitive since one would expect 
significant information loss during the reduction to a graph.
A careful look at the algorithm presented in the next section would reveal that this
is not the case. 
While most information in modes $3,\ldots, m$ are lost, one can still estimate
$Z$ from the first two modes, which suffices for the purpose of detecting
planted partitions.
This observation is reinforced by the recent study of~\citet{Florescu_2016_conf_COLT}, where the authors show that a reduction based spectral approach can optimally detect two partitions all the way down to the limit of identifiability of the partitions.

We conclude this section with a brief mention of the wide variety of other
higher order learning methods, which include 
tensor based clustering algorithms~\citep{Shashua_2006_conf_ECCV,Chen_2009_jour_JourCV,AriasCastro_2011_jour_EJStat,Ochs_2012_conf_CVPR},
other unsupervised tensorial learning schemes~\citep{Duchenne_2011_jour_TPAMI,Nguyen_2015_conf_CVPR},
as well as related optimisation approaches~\citep{Leordeanu_2012_conf_AISTATS,RotaBulo_2013_jour_TPAMI,Jain_2013_conf_ICCV},
and can easily be represented as instances of the uniform hypergraph partitioning problem.
In~\citet{Ghoshdastidar_2015_conf_ICML}, we showed that 
most of the above methods can be unified by a general
\emph{tensor trace maximisation} (TTM) problem.
A spectral solution to this problem is analysed in the present paper,
and thus, we believe that some of our conclusions can also be extended
to these alternative approaches.
We also numerically compare with some of these methods in Section~\ref{sec_experiment}.

\section{Tensor Trace Maximisation (TTM)}
\label{sec_ttm}

In this section, we briefly recap the work in~\citet{Ghoshdastidar_2015_conf_ICML} and then present the basic approach
that will be later analysed and modified in the remainder of the paper.
Let $(\Vc,\Ec,w)$ be a given weighted uniform hypergraph,
where $\Ec$ is the collection of all sets of $m$ vertices, and $w:\Ec\to[0,1]$ associates a weight with every edge.
We consider the problem of partitioning $\Vc$ into $k$ disjoint sets, $\Vc_1,...,\Vc_k$,
such that the total weight of edges within each cluster is high, 
and the partition is `balanced'.
In the case of graphs, \ie for $m=2$, 
the popular heuristic for achieving these two conditions
is the normalised cut minimisation problem, or 
equivalently normalised associativity maximisation problem~\citep{Shi_2000_jour_TPAMI,vonLuxburg_2007_jour_StatComp}.
We extend this approach to uniform hypergraphs.

\subsection{TTM Approach and Algorithm}

We consider the problem of finding the partition of the vertices that maximises the normalised associativity.
This is subsequently
formulated in terms of a tensor trace maximisation objective.
We define few terms.
The degree of any node $v\in\Vc$ is the total weight of edges on which
$v$ is incident, \ie $\degr(v) = \sum\limits_{e\in\Ec: v\in e} w_e$.
For any collection of nodes $\Vc_1\subseteq\Vc$, 
we define its volume as
$\vol(\Vc_1) = \sum\limits_{v\in\Vc_1} \degr(v)$ 
and its associativity as 
$\assoc(\Vc_1) = \sum\limits_{e\in\Ec: e\subset\Vc_1} w_e$,
which is the total weight of edges contained within $\Vc_1$.
The normalised associativity 
of a partition $\Vc_1,...,\Vc_k$ is given as
\begin{equation}
 \textup{N-Assoc}(\Vc_1,\ldots,\Vc_k) = \sum_{i=1}^{k} 
 \frac{\assoc(\Vc_i)}{\vol(\Vc_i)} \;.
 \label{eq_partition_assoc}
\end{equation}
Observe that the above definitions coincide with 
the corresponding terms in graph literature~\citep{Shi_2000_jour_TPAMI}.
We now follow the popular goal of finding clusters that
maximises the normalised associativity~\eqref{eq_partition_assoc}.
In the case of graphs, it is well known that the problem can be reformulated
in terms of the adjacency matrix of the graphs, which results
in a matrix trace maximisation problem~\citep{vonLuxburg_2007_jour_StatComp}.
Furthermore, a spectral relaxation allows one to find an approximate
solution for the problem by computing the $k$ dominant eigenvectors
of the normalised adjacency matrix.\footnote{
Dominant eigenvectors are the orthonormal eigenvectors corresponding to the largest $k$ eigenvalues.}
A similar approach is possible in the case of uniform hypergraphs.
Let $\Ab$ be the  adjacency tensor (of order $m$), \ie
\begin{align}
 \Ab_{i_1i_2\ldots i_m} = \left\{ \begin{array}{ll}
                w_{\{i_1,i_2,\ldots,i_m\}}  & \text{if } i_1,i_2,\ldots,i_m \text{ are distinct},
                \\
                0    & \text{otherwise}.
               \end{array}\right.
\end{align}
Define $\beta_1,...,\beta_m\in[0,1]$ with $\sum\limits_{l=1}^m \beta_l = 1$,
and $Y^{(1)},\ldots, Y^{(m)} \in\R^{n\times k}$ with
$Y_{ij}^{(l)} =  \left(\frac{\1\{i\in\Vc_j\}}{\vol(\Vc_j)}\right)^{\beta_l}$.
Then, one can rewrite~\eqref{eq_partition_assoc} (see appendix for details) as
\begin{equation}
 \textup{N-Assoc}(\Vc_1,\ldots,\Vc_k) = \frac{1}{m!}
 \textup{Trace}\left(\Ab \times_1 {Y^{(1)}}^T \times_2 {Y^{(2)}}^T
 \times_3 \ldots \times_m {Y^{(m)}}^T\right)\;,
 \label{eq_partition_assoc_tensor}
\end{equation}
where $\times_l$ denotes the mode-$l$ product.
Thus, for some chosen parameters $\beta_1,\ldots,\beta_m$,
one can pose the associativity maximisation problem as a tensor
trace maximisation (TTM).

In~\citet{Ghoshdastidar_2015_conf_ICML}, we showed that the above optimisation has connections with the
tensor eigenvalue problem~\citep{Lim_2005_conf_CAMSAP} and the tensor diagonalisation problem~\citep{Comon_2001_conf_ICISP}.
More interestingly, it also
lies at the heart of several higher-order learning algorithms.
For instance, $\beta_1=\ldots=\beta_m=\frac{1}{m}$ results in the method of~\citet{Shashua_2006_conf_ECCV}, while the same strategy when used with $k=1$
has been used to successively extract clusters~\citep{RotaBulo_2013_jour_TPAMI,Leordeanu_2012_conf_AISTATS}.
A similar idea lies in some tensor matching algorithms~\citep{Duchenne_2011_jour_TPAMI,Nguyen_2015_conf_CVPR}.
Another strategy is to set $\beta_1=\beta_2=\frac12$ and
$\beta_3=\ldots=\beta_m=0$.
This squeezes the tensor into a matrix, which allows one to use subsequently graph partitioning tools.
Our algorithm described below takes this route, and similar ideas have been previously used~\citep{Agarwal_2005_conf_CVPR,AriasCastro_2011_jour_EJStat}.
This reduction also corresponds to the clique expansion of a hypergraph,
where every $m$-way edge is replaced by $\binom{m}{2}$ pairwise edges.

We list our basic spectral approach in Algorithm~\ref{alg_ttm},
and we later study the consistency of TTM and its sampled variants.
Note that a spectral relaxation of the problem is a two-fold procedure,
where first we construct a matrix from the affinity tensor $\Ab$,
and then relax the problem into a matrix spectral decomposition type objective.
This principle is reminiscent of the classical technique for studying
spectral properties of hypergraphs~\citep{Bolla_1993_jour_DiscreteMath},
and is closely related to approach of the clustering
graph approximations of hypergraphs~\citep{Agarwal_2006_conf_ICML}.

\begin{varalgorithm}{TTM}
\caption {: Spectral relaxation of tensor trace maximisation problem}
\label{alg_ttm}
\begin{algorithmic}[1]
 \REQUIRE Affinity tensor $\Ab$ of the $m$-uniform hypergraph $(\Vc,\Ec, w)$, where $|\Vc|=n$.\\
 \STATE Read $\Ab$, and compute the $n\times n$ matrix $A$ as
   $A_{ij} = \sum\limits_{i_3,\ldots,i_m=1}^n \Ab_{ij i_3\ldots i_m}$.
 \STATE Let $D\in\R^{n\times n}$ be diagonal with $D_{ii} = \sum\limits_{j=1}^n A_{ij}$,
   and $L = D^{-1/2}AD^{-1/2}$.
 \STATE Compute $k$ dominant eigenvectors of $L$, denoted by $X\in\R^{n\times k}$ .
 \STATE Normalise rows of $X$ to have unit norm, and denote this matrix as $\overline{X}$.\\
 \STATE Run $k$-means on the rows of $\overline{X}$.\\
 \ENSURE Partition of $\Vc$ that correspond to the clusters obtained from $k$-means.\\
\end{algorithmic}
\end{varalgorithm}

A careful look at computational complexity of the algorithm will be helpful
for our discussions in Section~\ref{sec_Tetris}.
To this end, we note that our analysis assumes the use 
$k$-means approach of~\citet{Ostrovsky_2012_jour_JACM}, which
has a complexity of $O(k^2n+k^4)$ since the data is embedded in a $k$-dimensional space.
Furthermore, Steps~2 to~4 involve only matrix operations with the eigenvector computation 
being the most expensive operation. One may compute the $k$ dominant eigenvectors 
using power iterations, which can be done provably in $O(k n^2\ln(k n))$ runtime~\citep{Boutsidis_2015_conf_ICML}.
However, the computational bottleneck of the algorithm is Step~1, which has complexity of $m^2|\Ec| = O(m^2n^m)$.
This is not surprising since any network partitioning method should have 
have complexity at least linear in the number of edges.
But it gets quite challenging in computer vision problems,
where one often requires to consider higher order relations, for instance
$m=4$ used in~\citet{Duchenne_2011_jour_TPAMI}, $m=5$ in~\citet{Chen_2009_jour_JourCV} and even $m=8$ in~\citet{Govindu_2005_conf_CVPR}.
The aim of Section~\ref{sec_Tetris} is to reduce this complexity to
$m^2N$ by sampling only $N\ll n^m$ edges.

\section{Consistency of Algorithm~\ref{alg_ttm} under Planted Partition Model}
\label{sec_ttm_consistency}

The first task in our agenda is to analyse the basic TTM approach.
This section extends the results in~\citet{Ghoshdastidar_2015_conf_ICML}
to the the planted partition model for sparse weighted hypergraphs
described in Section~\ref{sec_model}.
Recall that our aim is to derive an upper bound on $\err$,
where $\psi$ and $\psi'$ denote the true and the estimated clusters. 
Moreover, for the purpose of analysis we assume that the $k$-means
step is performed using Lloyd's approach with the seeding
described in~\citep{Ostrovsky_2012_jour_JACM}. The reason for this consideration
is the known theoretical guarantee for this method.

We briefly recall the planted model for weighted $m$-uniform hypergraphs described in Section~\ref{sec_model}.
An underlying function $\psi$ groups the $n$ nodes into $k$ clusters, and $\psi_i$ denotes the true cluster of node $i$. 
For any edge $e=\{i_1,i_2,\ldots, i_m\}$, its weight $w_e$ is a random variable taking values in $[0,1]$ with mean given by~\eqref{eq_model}, where the parameters $\alpha_{n}\in[0,1]$ and symmetric $m$-way tensor $\Bb \in [0,1]^{k \times k \times \ldots \times k}$, respectively, govern the mean edge weight and the relative weights of edges formed among nodes from different classes.\footnote{Note that $\alpha_n$ plays the role of a sparsity parameter commonly introduced to define sparse stochastic block models~\citep{Lei_2015_jour_AnnStat}, and a smaller $\alpha_n$ increases the complexity of the problem. However, unlike the unweighted case, all edges are present in our setting but most of the weights are very small.}
For instance, in Section~\ref{sec_ttm_consistency_example}, we consider an example where
given $p, q\in[0,1]$, the tensor $\Bb$ is constructed such that $\E[w_e] = \alpha_n(p+q)$ if all nodes in the edge belong to the same cluster, and  $\E[w_e] = \alpha_n q$ if the participating nodes are from different clusters. Thus, in this case, the mean weight of edges residing within in each cluster is larger than inter-cluster edge weights.
In addition to above, we also assume that all edge weights $(w_e)_{e\in\Ec}$ are mutually independent.

Consider a random $m$-uniform hypergraph $(\Vc,\Ec, w)$ generated according to the above model.
As a consequence of~\eqref{eq_model}, the expected affinity tensor of the hypergraph
\begin{equation}
\E_H[\Ab_{i_1i_2\ldots i_m}] = \left\{ \begin{array}{ll}
\alpha_n \Bb_{\psi_{i_1}\psi_{i_2}\ldots\psi_{i_m}} & \text{if } i_1,i_2,\ldots,i_m \text{ are distinct, and}\\
0 & \text{otherwise,}\end{array}\right.
\label{eq_exp_affi}
\end{equation}
has a block structure, ignoring entries with repeated indices. Obviously, the $k^m$ blocks are aligned with the underlying clusters, which gives rise to the representation mentioned in~\eqref{eq_A_hosvd_form}.
Algorithm~\ref{alg_ttm} first squeezes the adjacency tensor $\Ab$ to a $n\times n$ matrix $A$.  
To analyse the algorithm in the expected case, let $\Ac = \E_H[A]$ and $\Dc = \E_H[D]$,
where $A$ and $D$ are the matrices computed in Algorithm~\ref{alg_ttm}.
Observe that if  the algorithm had access to the expected affinity tensor~\eqref{eq_exp_affi},
then $\Ac$ corresponds to the matrix computed in the first step of the algorithm,
and $\Dc_{ii} = \sum\limits_{j=1}^n \Ac_{ij}$.
From the definition of the model, it can be seen that
$\Ac_{ii}=0$ for all $i$, and for $i\neq j$,
\begin{align}
 \Ac_{ij} 
 = (m-2)!\sum_{\substack{i_3<i_4<\ldots<i_m,\\i,j\notin\{i_3,\ldots,i_m\}}} 
 \alpha_{n} \Bb_{\psi_i\psi_j\psi_{i_3}\ldots\psi_{i_m}} \;,
 \label{eq_Ac_expn}
\end{align}
where the factor $(m-2)!$ takes into account all permutations of $\{i_3,\ldots, i_m\}$.
The key observation here is that $\Ac_{i j} = \Ac_{i'j'}$ whenever
$\psi_i=\psi_{i'}$ and $\psi_j=\psi_{j'}$,
which holds since, under the present model, nodes in the same cluster are statistically identical.
Thus, one can define a matrix $G\in\R^{k\times k}$ such that 
$\Ac_{ij} = G_{\psi_i\psi_j}$ for all $i\neq j$.
This implies that, ignoring the diagonal entries, $\Ac$ is essentially of rank $k$.

Let $Z\in\{0,1\}^{n\times k}$ be the assignment matrix 
corresponding to partition $\psi$, \ie $Z_{ij}=\1\{i\in\psi(j)\}$,
and let the sizes of the $k$ clusters be $n_1\geq n_2 \geq\ldots \geq n_{k}$.
We define
\begin{equation}
 \delta = \lambda_{k}(G)\min_{1\leq i\leq n} 
\frac{n_{\psi_i}}{\Dc_{ii}} - 
\max_{1\leq i,j\leq n} \left|\frac{G_{\psi_i\psi_i}}{\Dc_{ii}} 
- \frac{G_{\psi_j\psi_j}}{\Dc_{j j}}\right|,
\label{eq_delta_charac}
\end{equation}
where $\lambda_k(G)$ is the smallest eigenvalue of $G$.
The following lemma, proved in the appendix, shows that if $\delta>0$, then Algorithm~\ref{alg_ttm} correctly identifies the underlying clusters in the expected case.
\begin{lemma}
 \label{lem_ttm_expected}
Let $\Lc = \Dc^{-1/2}\Ac\Dc^{-1/2}$.
  If $\delta$ in~\eqref{eq_delta_charac} satisfies $\delta>0$, then
  there exists an orthonormal matrix $U\in\R^{k\times k}$ such that
  the $k$ leading orthonormal eigenvectors of $\Lc$ correspond to the columns of the 
  matrix  $\Xc = Z(Z^TZ)^{-1/2}U$.
\end{lemma}
It is easy to see that $\Xc$ has $k$ distinct rows,
each corresponding to a true cluster. 
Hence, clustering the rows of $\Xc$ (or its row normalised form) 
using $k$-means gives an accurate clustering of the nodes.
In the random case, however, the dominant eigenvectors of $L$ computed in \ref{alg_ttm} need not always reflect the true assignment matrix $Z$.
The following result shows that under certain conditions on the model parameters, the eigenvectors are still close to $\Xc$, and hence, the number of mis-clustered nodes~\eqref{eq_error} grows slowly. 
\begin{theorem}
 \label{thm_ttm_consistency}
 Let $(\Vc,\Ec, w)$ be a random $m$-uniform hypergraph on $|\Vc|=n$ vertices
 generated from the model described above.
Define $d = \min\limits_{1\leq i \leq n} \E_H [\deg(i)]$
and, without loss of generality, assume that the cluster sizes are $n_1\geq n_2\geq \ldots \geq n_{k}$.
Let $\delta$ be as defined in~\eqref{eq_delta_charac}.
\\
There exists an absolute constant $C>0$, such that, if $\delta>0$ and
\begin{equation}
\delta^2 d>\frac{C k n_1 (\ln n)^2}{n_k}
\label{eq_dn_condns}
\end{equation}
for all large $n$, then 
with probability $\left(1-o(1)\right)$, the partitioning error for TTM is
\begin{equation}
\err = O\left(\frac{k n_1 \ln n}{\delta^2d}\right).
\label{eq_ttm_error_bound}
\end{equation}
\end{theorem}
The bound in~\eqref{eq_ttm_error_bound} along with the condition in~\eqref{eq_dn_condns} immediately suggests that TTM is weakly consistent, \ie $\err = o(n)$ or the fractional of mis-clustered vertices vanishes as $n\to\infty$.
However, in certain (dense) cases, even $\err\to0$ as we will discuss later.
Note that $d$ grows with $n$ though this dependence is not made explicit in the notation. We also allow $k$ to vary with $n$. 
The condition on $\delta^2d$ in~\eqref{eq_dn_condns} ensures that the hypergraph is
sufficiently dense so that the following three conditions hold, respectively:
(i) the matrix $A$ computed in Algorithm~\ref{alg_ttm} concentrates near its expectation,
(ii) the $k$ dominant eigenvectors of $L$ contain information about the partition, and
(iii) the $k$-means step provides a near optimal solution.   
While restricting $d$ from below in~\eqref{eq_dn_condns}
essentially limits the sparsity of the hypergraph, the quantity $\delta$ on the
other hand quantifies the complexity of the model.

The threshold for identifying two partitions, derived in~\citet{Florescu_2016_conf_COLT},  shows that the condition in~\eqref{eq_dn_condns} differs from the threshold for identifiability only by logarithmic factors. 
These extra $\ln n$ factors arise since (i) we consider weighted hypergraphs, and (ii) we do not substitute $k$-means by alternative strategies.
Note that even  works on stochastic block model~\citep{Lei_2015_jour_AnnStat,Gao_2015_arxiv_03772} do not consider these two factors, and hence, even in the case of graphs, it is not known till date whether the logarithmic terms can be avoided when these practical aspects are included in the analysis.
We point out that the present analysis incorporates the guarantees for $k$-means derived in~\citet{Ostrovsky_2012_jour_JACM}, which was also used in our earlier work~\citep[see Lemma 4.8 of][]{Ghoshdastidar_2017_jour_AnnStat}.

\subsection{A Special Case}
\label{sec_ttm_consistency_example}
To gain insights into the implications of Theorem~\ref{thm_ttm_consistency},
we consider the following special case of the planted partition model.
The partition $\psi$ is defined such that the $k$ clusters are of equal size.
Moreover, the tensor $\Bb$ in~\eqref{eq_model} is given by
$\Bb_{j_1 j_2 \ldots j_m} = (p+q)$ if $j_1=j_2=\ldots=j_m$, and $q$ otherwise,
where $p, q\in[0,1]$ with $q\leq(1-p)$.
Thus, in this model, edges residing within each cluster
have a high weight (in the expected sense) as compared to other edges.\footnote{%
The model considered here may be viewed as the four parameter stochastic 
block model~\citep{Rohe_2011_jour_AnnStat} defined by the parameters
$(n, k, p_n, q_n)$, where $n$ nodes are divided into $k$ partitions of equal size.
Edges within a cluster occur with probability $(p_n+q_n)$, while inter-cluster edges occur with 
probability $q_n$. Here, we set $p_n=\alpha_n p$ and $q_n = \alpha_n q$ for some constants
$p, q\in[0,1]$ with $q\leq(1-p)$.}
This model corresponds to the decomposition of $\Ab$ mentioned in~\eqref{eq_A_cp_form}.
We state the following consistency result for dense hypergraphs.
\begin{corollary}
 \label{cor_ttm_consistency_balanced_dense}
 Let $\alpha_n=1$ and $k = O\left(\frac{n^{1/4}}{\ln n}\right)$. Then with probability $(1-o(1))$,
 \begin{equation}
 \err = O\left( \frac{n^{(3-m)/2}}{(\ln n)^{2m-3}} \right)\;.
 \end{equation}
 \end{corollary}
 According to the notions of consistency defined in~\citet{Mossel_2013_arxiv}, it can be seen
 that for $m=2$, Algorithm~\ref{alg_ttm} is {weakly consistent}, \ie $\err = o(n)$.
 We note here that, in this sense, the algorithm is not worse than  spectral clustering 
 that is also known to be weakly consistent~\citep{Rohe_2011_jour_AnnStat}.
 However, for $m\geq3$, $\err = o(1)$ for Algorithm~\ref{alg_ttm}, which
 implies that it is {strongly consistent} in this case.
 In other words, the algorithm can exactly recover the partitions for large $n$.
 This conclusion is intuitively acceptable since in this case, uniform hypergraphs
 for large $m$ have a large number of edges that provides `more' information
 about the partition, providing a smaller error rate.

In the sparse regime, the question one is interested in is the minimum level of sparsity under which
weak consistency of an algorithm can be proved. The following result answers this question.  For the case of graphs $(m=2)$,~\citet{Lei_2015_jour_AnnStat} showed that weak consistency is achieved by spectral clustering for $\alpha_n \geq \frac{C\ln n}{n}$, which matches our result upto a factor of $(\ln n)^2$.
In fact, our proof also allows the difference to be reduced to a factor of $\omega(\ln n)$, but this difference has negligible effect in practice.
\begin{corollary}
 \label{cor_ttm_consistency_balanced_sparse}
 Let $k = O\left({\ln n}\right)$. There exists an absolute constant $C>0$, such that, if 
 \begin{equation}
 \alpha_n \geq \frac{C (\ln n)^{2m+1}}{n^{m-1}} \;,
 \label{eq_sparsity_lowerbound}
 \end{equation}
 then $\err = O\left( \frac{n}{(\ln n)^2} \right) = o(n)$ with probability $(1-o(1))$.
 \end{corollary}

While the stochastic block model has been extensively studied for graphs, the existing hypergraph literature provides consistency results for only two other approaches:
\begin{itemize}
\item
 a uniform hypergraph partitioning method that uses a higher order singular value 
decomposition (HOSVD) of the adjacency tensor~\citep{Govindu_2005_conf_CVPR,Ghoshdastidar_2014_conf_NIPS}, and
\item
 a non-uniform hypergraph partitioning approach that solves a spectral relaxation of the 
 normalised hypergraph cut (NH-Cut) problem~\citep{Zhou_2007_conf_NIPS,Ghoshdastidar_2017_jour_AnnStat}.
 \citet{Florescu_2016_conf_COLT} also use a similar method.
\end{itemize}
We comment on the theoretical performance of TTM in comparison with these two approaches.
In particular, we focus on the settings of Corollaries~\ref{cor_ttm_consistency_balanced_dense} and~\ref{cor_ttm_consistency_balanced_sparse}.
The following remark is quite surprising since both TTM and NH-Cut reduce
the hypergraphs to graphs, and hence, apparently incur some loss. Yet both
outperform HOSVD, which appears to be the most natural solution according 
to the representation in~\eqref{eq_A_hosvd_form}. This fact is also validated numerically in Section~\ref{sec_spec_comp}.
\begin{remark}
\label{rem_spectral_comp}
Under the setting of Corollary~\ref{cor_ttm_consistency_balanced_dense}, the error
bound for the NH-Cut algorithm is 
\begin{displaymath}
\textup{Err}_\textup{NH-Cut}(\psi,\psi') = O\left(\frac{n^{(3-m)/2}}{(\ln n)^{2m-3}}\right)
\end{displaymath}
with probability $(1-o(1))$, while the corresponding bound for HOSVD algorithm is
\begin{displaymath}
\textup{Err}_\textup{HOSVD}(\psi,\psi') = O\left(\frac{n^{(4-m)/2}}{(\ln n)^{2m-1}}\right)\;.
\end{displaymath}
Thus the performance of NH-Cut is similar to TTM, and both methods have a smaller error 
bound than HOSVD.

Similarly, in the case of Corollary~\ref{cor_ttm_consistency_balanced_sparse},
the lower bound on sparsity for NH-Cut is same as in~\eqref{eq_sparsity_lowerbound}
up to a constant scaling. 
However, HOSVD achieves weak consistency  only for 
\begin{displaymath}
 \alpha_n \geq \frac{C' (\ln n)^{m+1.5}}{n^{(m-1)/2}} 
\end{displaymath}
for some $C'>0$. This is larger than the allowable sparsity for TTM or NH-Cut.
\end{remark}

\subsection{Proof of Theorem~\ref{thm_ttm_consistency}}
Here, we give an outline of the proof of Theorem~\ref{thm_ttm_consistency}
using a series of technical lemmas. The proofs of these results
are given in the appendix.
The proof has a modular structure which consists of 
(i) deriving certain conditions on the model parameters
such that Algorithm~\ref{alg_ttm} incurs no error in the expected case,
(ii) subsequent use of
matrix concentration inequalities and spectral perturbation bounds
to claim that (almost surely) the dominant eigenvectors in the random case do not  
deviate much from the expected case, and
(iii) finally, the proof of correctness of the $k$-means step.

Recall that the first step of the proof is taken care of by Lemma~\ref{lem_ttm_expected}.
For convenience, define $\Dmin = \min\limits_{1\leq i \leq n} \Dc_{ii}$.
One can see that $\Dmin = (m-1)!d$. Hence, for the 
subsequent analysis as well as for all the proofs, it is more convenient 
to expand~\eqref{eq_dn_condns} as 
\begin{equation}
\Dmin>\max\left\{C_1\ln n, \frac{C_2\ln n}{\delta^2}, \frac{C_3 k n_1 \ln n}{n_k \epsilon^2 \delta^2}\right\}\;,
\label{eq_Dmin_condns}
\end{equation}
where the constants $C_1,C_2,C_3>0$ take into account the additional factor of $(m-1)!$.
Note that the last term dominates, and corresponds to~\eqref{eq_dn_condns} when $\epsilon=(\ln n)^{-1/2}$.
The subsequent results show that the eigenvector matrix $X$
computed from a random realisation of the hypergraph is 
close to $\Xc$ almost surely, and hence,
one can expect a good clustering in the random case.

Lemma~\ref{lem_ttm_random_deviation} proves a concentration bound for 
the normalised affinity matrix $L$ computed in Algorithm~\ref{alg_ttm}.
The proof, given in the appendix, relies on an useful characterisation of the matrix $A$.
To describe this representation, we define for each edge 
$e\in\Ec$, a matrix $R_e \in\{0,1\}^{n\times n}$ as
$(R_e)_{ij} = 1$ if $i, j\in e, i\neq j$, and zero otherwise.
Quite similar to the representation of~\eqref{eq_Ac_expn}, one can note that
\begin{equation}
A = (m-2)! \sum_{e\in\Ec} w_eR_e \;.
\label{eq_A_charac}
\end{equation}
This characterisation is quite useful since the independence of $(w_e)_{e\in\Ec}$
ensures that $A$ is represented as a sum of independent random matrices, and hence,
one can use matrix concentration inequalities~\citep{Tropp_2012_jour_FOCM}
 to derive a tail bound for $\Vert A - \Ac\Vert_2$.
\begin{lemma}
\label{lem_ttm_random_deviation}
If there exists $n_0$ such that $\Dmin>9(m-1)!\ln n$ for all $n\geq n_0$, then
with probability $\left(1-O(n^{-2})\right)$,
\begin{equation}
\Vert L - \Lc \Vert_2 \leq 12\sqrt{\frac{(m-1)!\ln n}{\Dmin}}\;.
\end{equation}
\end{lemma}
The above result directly leads to a bound on the perturbation of the eigenvectors
as shown in Lemma 4.7 of~\citet{Ghoshdastidar_2017_jour_AnnStat}.
The result adapted to our setting is stated below.
\begin{lemma}
\label{lem_ttm_perturbation}
Assume there is an $n_0$ such that $\Dmin>9(m-1)!\ln n$ and
$\delta > 24\sqrt{\frac{(m-1)!\ln n}{\Dmin}}$ for all $n\geq n_0$.
Then the following statements hold with probability $\left(1-O(n^{-2})\right)$.
\begin{enumerate}
 \item The matrix $X$ does not have any row with zero norm,
     and hence, its row normalised form, denoted by $\overline{X}$, is well-defined.
 \item  There is an orthonormal matrix $Q\in\R^{k \times k}$ such that
\begin{equation}
\left\Vert \overline{X} - ZQ\right\Vert_F \leq 
\frac{24}{\delta}\sqrt{\frac{(m-1)!2k n_1 \ln n}{\Dmin}}\;.
\label{eq_perturbation_bound}
\end{equation}
\end{enumerate}
\end{lemma}

Finally, we analyse the $k$-means step of the algorithm, where the rows of $\overline{X}$
are assigned to $k$ centres. Define $S\in\R^{n\times k}$ such that $S_{i\cdot}$
denotes the centre to which $\overline{X}_{i\cdot}$ is assigned. Also define the collection of
nodes $\Vc_{err} \subset \Vc$ such that
\begin{equation}
 \Vc_{err} = \left\{ i\in\Vc : \Vert S_{i\cdot} - Z_{i\cdot} Q\Vert_2 \geq \frac{1}{\sqrt{2}}\right\}\;.
\label{eq_error_charac}
\end{equation}
The following result, adapted from~\citet{Ghoshdastidar_2017_jour_AnnStat}, shows that on one hand $\Vc_{err}$ contains all the mis-labelled nodes,
whereas, on the other, it proves that under the conditions of Theorem~\ref{thm_ttm_consistency},
the $k$-means algorithm of~\citet{Ostrovsky_2012_jour_JACM} finds a near optimal solution
for which $|\Vc_{err}|$ can be bounded from above. 
\begin{lemma}
\label{lem_ttm_kmeans}
 Under the conditions stated in~\eqref{eq_Dmin_condns}, for small enough $\epsilon$,
\begin{equation}
   \err \leq |\Vc_{err}| \leq 8(1+\epsilon^2)^2 \Vert \overline{X} - ZQ\Vert_F^2
\label{eq_error_bound}
\end{equation}
with probability $\left(1-O(n^{-2}+\sqrt{\epsilon})\right)$.
\end{lemma}
Theorem~\ref{thm_ttm_consistency} follows by setting $\epsilon=(\ln n)^{-1/2}$, and using the bound
on $\Vert \overline{X} - ZQ\Vert_F$ in~\eqref{eq_perturbation_bound}.

\section{Sampling Techniques for Algorithm~\ref{alg_ttm}}
\label{sec_Tetris}

We now present the second, and key, contribution in this work.
Recall from the discussions in Section~\ref{sec_ttm} that the overall computational complexity of TTM is  $O(m^2n^m +k n^2\ln(k n) +k^2n +k^4)$,
where the first term clearly dominates for $m\geq3$.
A practical solution to this problem is to simply compute the weights of few edges, or equivalently, sample few entries of the adjacency tensor $\Ab$.
This strategy has often been used in computer vision, but to the best of our knowledge, there is no known theoretical study of the approach.
The only relevant theoretical works~\citep{Bhojanapalli_2015_arxiv_05023,Jain_2014_conf_NIPS} are in a different context,
where the authors study factorisation of partially observed tensors.
While the latter work assumes an uniform sampling, the former presents distributions that are more adapted to the tensor.
We later compare these results with our findings.

In contrast, practical higher order learning methods exhibit considerable variety in sampling techniques. \citet{Govindu_2005_conf_CVPR} used a sampling that uniformly selects fibers of the tensor, which is similar in spirit to the well known column sampling technique for matrices.
Ideas along the same lines, and also a Nystr{\"o}m approximation, for the HOSVD based approach were suggested in~\citet{Ghoshdastidar_2015_conf_AAAI}.
A more efficient technique of iterating between sampling and clustering was used in~\citet{Jain_2013_conf_ICCV} and~\citet{Chen_2009_jour_JourCV}, where one starts with a naive sampling to get approximate partitions and then iteratively improves the result by sampling edges aligned with partitions.
Matching algorithms~\citep{Duchenne_2011_jour_TPAMI} exploit side information to prioritise edges with significant weights. Other heuristics have also been suggested in some works.

We formally study the following problem. 
Suppose we are given a certain distribution $(p_e)_{e\in\Ec}$ on the set of all $m$-way edges $\Ec$. Let $N$ edges be sampled with replacement according to the given distribution. 
We aim to find the minimum sample size required such that corresponding partitioning algorithm (with edge sampling) is still weakly consistent.
In this paper, we assume that the core partitioning approach is TTM,
and the sampling only affects Step 1 of the algorithm, where we replace $A$
by its sample estimate, denoted by $\Ah$.
A requirement of the estimator should be its unbiasedness, \ie
$\E_{S|H}[\Ah] = A$, where the expectation is with respect to the sampling distribution given an instance of the random hypergraph. 
Based on~\eqref{eq_A_charac}, we propose to use an unbiased estimator of the form
\begin{equation}
 \Ah = \frac{(m-2)!}{N} \sum_{e\in\Ic} \frac{w_e}{p_e} R_e \;,
 \label{eq_Ah_defn}
\end{equation}
where $\Ic\subset\Ec$ with $|\Ic|=N$  is the collection of sampled edges (with possible duplicates).
The matrix $R_e \in\{0,1\}^{n\times n}$ is such that
$(R_e)_{i j} = 1$ if $i, j\in e, i\neq j$, and zero otherwise.
The overall method is listed below, and one can easily see that its runtime is
$O(m^2N +kn^2\ln(kn) +k^2n +k^4)$.

\begin{varalgorithm}{Sampled TTM}
\caption {: TTM where a sampled set of edge weights are observed}
\label{alg_sampledttm}
\begin{algorithmic}[1]
 \REQUIRE Distribution $(p_e)_{e\in\Ec}$ on the set of all edges $\Ec$;
 \newline
 Affinity tensor $\Ab$, which is not observed, but requested entries can be observed.\\
 \STATE Sample a collection of $N$ edges $\Ic\in\Ec$ with replacement.
 \STATE Observe entries of $\Ab$ corresponding to edges in $\Ic$, and compute $\Ah$ using~\eqref{eq_Ah_defn}.
 \STATE Run Steps 2-5 of Algorithm~\ref{alg_ttm} using $\Ah$ instead of $A$.\\
 \ENSURE Partition of $\Vc$ that correspond to the clusters obtained from $k$-means.\\
\end{algorithmic}
\end{varalgorithm}

\subsection{Consistency of Sampled Variants of Algorithm~\ref{alg_ttm}}

We now analyse the performance of Algorithm~\ref{alg_sampledttm} for any given edge sampling distribution $(p_e)_{e\in\Ec}$.
The main message of the following result is that the algorithm remains weakly consistent even if we use very small number of sampled edges, \ie $N=o(n^m)$.
However, the minimum sample size required depends on the sampling distribution. 

\begin{theorem}
 \label{thm_ttm_consistency_sampled}
Let $N$ edges be sampled with replacement according to probability distribution $(p_e)_{e\in\Ec}$,
and let $\beta>0$ be such that $\P_H\left(\max\limits_{e\in\Ec} \frac{w_e}{p_e} >\beta\right) = o(1)$.
Let $\delta$ be as defined in~\eqref{eq_delta_charac}.
There exist absolute constants $C,C'>0$, such that, if $\delta>0$,
\begin{equation}
\delta^2d>C\frac{k n_1 (\ln n)^2}{n_k}
\quad\text{ and }\quad
N >C'\left(1+\frac{2\beta}{d}\right)\frac{k n_1 (\ln n)^2}{n_k \delta^2}
\label{eq_dn_condns_sampled}
\end{equation}
for all large $n$, then 
with probability $\left(1-o(1)\right)$, 
\begin{equation}
\err = O\left(\frac{k n_1\ln n}{\delta^2}\left(
\frac{1}{d} + \frac{1}{N}+\frac{2\beta}{Nd}\right)\right) = o(n).
\label{eq_ttm_error_bound_sampled}
\end{equation}
\end{theorem}
Note that the above probability is with respect to both the randomness of the hypergraph
and edge sampling. 
The above result is similar to Theorem~\ref{thm_ttm_consistency} except for  the additional condition and error term associated with the number of sampled edges $N$.
We mention here that the constant $C$ in~\eqref{eq_dn_condns_sampled} is different from the one used in Theorem~\ref{thm_ttm_consistency}, but the quantity $\delta$ remains the same, and does not depend on the sampling distribution.
Since, the above error rate is $o(n)$, we can immediately conclude that 
the lower bound on $N$ in~\eqref{eq_dn_condns_sampled} is sufficient to ensure weak consistency of the sampled variant.
The result also shows that if we fix a particular sampling strategy, then
smaller $N$ is needed for denser and easier models (large $\delta^2d$).
This can be explained since for sparse hypergraphs, most edges have zero or negligibly small weights, and 
do not provide `sufficient information' about the true partition.

On the other hand, the sampling strategy plays a crucial role in the lower bound for $N$, but the dependence is only via the ratio of the edge weight to the sampling probability. We note that $\beta$ is a high probability upper
limit of this ratio,\footnote{The probability is with respect to the randomness of the planted model, and arises since the edge weights are random.}
and~\eqref{eq_dn_condns_sampled} suggests that a better sampling distribution is one for which $\beta$ is smaller. To clarify this observation, we state the result for two 
particular sampling distributions: (i) uniform sampling, and
(ii) sampling each edge $e$ with probability proportional to its weight, \ie
\begin{equation}
 p_e = \frac{w_e}{\sum\limits_{e'\in\Ec} w_{e'}} \qquad \text{ for all } e\in \Ec.
 \label{eq_ttm_wted_sampling}
\end{equation}
One can easily see that $\beta=\sum_e w_e$ in the latter case, whereas 
$\beta = \binom{n}{m}{\max_e w_e}$ for uniform sampling.
For ease of exposition, we restrict ourselves to the special case 
described in Section~\ref{sec_ttm_consistency_example}, and demonstrate the effect of these distributions on sample size.
\begin{corollary}
Consider the setting described in Section~\ref{sec_ttm_consistency_example}.
Define quantity $\xi$ such that $\xi=1$ for uniform sampling, and $\xi=\alpha_n$ for the 
weighted sampling of~\eqref{eq_ttm_wted_sampling}.
There exist constants $C,C'>0$, such that, if
\begin{equation}
\alpha_n>C\frac{k^{2m-1} (\ln n)^2}{n^{m-1}}
\quad\text{ and }\quad
N >C'\frac{\xi n k^{2m-1}(\ln n)^2}{\alpha_n} \;,
\label{eq_dn_condns_sampled_example}
\end{equation}
then $\err = o(n)$ with probability $\left(1-o(1)\right)$.
\label{cor_ttm_sampling_examples}
\end{corollary}

For simplicity, let us start with the case where $k=O(1)$. Then one has the lower bound $N = \Omega\left(\frac{n(\ln n)^2}{\alpha_n}\right)$ for uniform sampling, and $N = \Omega(n(\ln n)^2)$. 
Thus, the both sampling techniques have similar performance in the dense case $(\alpha_n=1)$, the gap between the lower bounds increase when $\alpha_n$ decays with $n$. In fact, in the most sparse setting possible in~\eqref{eq_dn_condns_sampled}, $\alpha_n = O(\frac{(\ln n)^2}{n^{m-1}})$ and
so, uniform sampling works only when one samples $\Omega(n^m)$ edges.\footnote{
Note that this observation is only true for weighted hypergraphs, where a small $\alpha_n$
implies that most edges have very small, but positive, weights. On the other hand, 
unweighted hypergraphs with small $\alpha_n$ implies that only few edges are present,
and sampling is not required if these edges are given a priori.} 
But with weighted sampling, one still needs only $\Omega(n(\ln n)^2)$ edges.

Possibility of achieving consistency with such a low sample size is quite remarkable, and has not been yet observed in any other tensor problem.
For instance, it may be argued that one can directly use the factorisation techniques for partially observed tensors studied in~\citet{Jain_2014_conf_NIPS} and~\citet{Bhojanapalli_2015_arxiv_05023}, to guarantee sampling rates derived in these works. 
We show here that this may not be a good strategy since such sampling can be often much larger than the rate derived in Corollary~\ref{cor_ttm_sampling_examples}.
We note here that the comparison is not entirely fair since, on one hand, we require only the clusters instead of the complete factorisation, whereas on the other hand, the results in related tensor sampling works are usually tied to an incoherence assumption that is violated in our setting.

For the comparison, we recall that in the setting of Corollary~\ref{cor_ttm_sampling_examples}, $\Ab$ has an approximate CP-decomposition of rank $(k+1)$ as shown in~\eqref{eq_A_cp_form}.
Furthermore, most works on tensors do not consider the case where entries decay with dimension, and so, we may assume $\alpha_n=1$.
The aforementioned works consider the problem of tensor factorisation, where the tensor is partly observed by means of some sampling.
\citet{Jain_2014_conf_NIPS} show that to obtain an accurate tensor factorisation, it is suffice to observe $\Omega(k^5 n^{m/2} (\ln n)^4)$ uniformly sampled entries.  
\citet{Bhojanapalli_2015_arxiv_05023} use a different sampling distribution, which essentially assigns more weight for larger entries quite similar to~\eqref{eq_ttm_wted_sampling}, and then prove a similar bound on the sample size (upto logarithmic factors).
The key difference of such bounds with Corollary~\ref{cor_ttm_sampling_examples}
is that the  $m$ in the exponential is tied to $n$ in other works, whereas it is tied to $k$ in our case---this improves efficiency significantly when $k$ grows much slower than $n$, which is clearly the case in clustering.

We elaborate on this further by considering the extreme values of $k$ and $\alpha_n$ possible in Corollaries~\ref{cor_ttm_consistency_balanced_dense} and~\ref{cor_ttm_consistency_balanced_sparse}, respectively.
First, let $\alpha_n=1$ and $k = O\left(\frac{n^{1/4}}{\ln n}\right)$, as in Corollary~\ref{cor_ttm_consistency_balanced_dense}.
Then both uniform and weighted sampling can guarantee weak consistency if 
$N = \Omega\left(n^{0.5m+0.75}(\ln n)^{3-2m}\right)$.
In contrast, the sample size from~\citep{Jain_2014_conf_NIPS} is $N=\Omega(n^{0.5m+1.25}(\ln n)^{-1})$, which is worse by a factor of about $\sqrt{n}$.
Turning to the setting of Corollary~\ref{cor_ttm_consistency_balanced_sparse}, we have $k=O(\ln n)$ and let $\alpha_n$ be at its lower bound in~\eqref{eq_sparsity_lowerbound}.
Then, the above result shows that while uniform sampling is poor, by sampling significant edges frequently, one needs only $N = \Omega\left(n(\ln n)^{2m+1}\right)$ edges for consistent partitioning.
In contrast, the weighted sampling of~\citet{Bhojanapalli_2015_arxiv_05023}
still needs $\Omega(n^{0.5m}(\ln n)^7)$ samples, which is much larger.

\subsection{Proof of Theorem~\ref{thm_ttm_consistency_sampled}}

We will further discuss the implications and limitations of weighted sampling, but first, we provide an outline for the proof of Theorem~\ref{thm_ttm_consistency_sampled}.
Recall that the sampled variant differs from core TTM algorithm only in the use of $\Ah$~\eqref{eq_Ah_defn} instead of $A$. 
Let us define $\Dh,\Lh$ for this case corresponding to $D,L$.
That is, $\Dh_{ii} = \sum_j \Ah_{ij}$ and $\Lh = \Dh^{-1/2}\Ah\Dh^{-1/2}$.
Note that $\Ah,\Dh$ are both unbiased estimates of $A,D$. 
The proof follows the lines of the proof of Theorem~\ref{thm_ttm_consistency}.
It is easy to see that the only difference is in Lemma~\ref{lem_ttm_random_deviation}, where instead of $\Vert L -\Lc\Vert_2$, we now need to compute a bound on
$\Vert\Lh - \Lc\Vert_2$. Observe that
\begin{equation*}
 \Vert \Lh - \Lc\Vert_2 \leq \Vert \Lh-L\Vert_2 + \Vert L -\Lc\Vert_2\;,
\end{equation*}
where the second term is bounded due to Lemma~\ref{lem_ttm_random_deviation}.
We have the following bound for the first part,
which can be derived using matrix Bernstein inequality.

\begin{lemma}
\label{lem_LLh_bound}
For large $n$ and under the conditions in~\eqref{eq_dn_condns_sampled}, with probability $1-o(1)$,
\begin{align}
 \Vert \Lh - L \Vert_2 
 &\leq 12\sqrt{\frac{\ln n}{N}\left(1+\frac{2\beta}{d}\right)}\;.
 \label{eq_LLh_bound}
\end{align}
\end{lemma}
This bound combined with Lemma~\ref{lem_ttm_random_deviation} implies
\begin{align}
 \Vert \Lh - \Lc \Vert_2 
\leq 12\sqrt{\frac{\ln n}{d}} + 
12\sqrt{\frac{\ln n}{N}\left(1+\frac{2\beta}{d}\right)} \;.
 \label{eq_LhLc_bound}
\end{align}
Let us denote the above upper bound by $\gamma_n$. Then one can restate
Lemmas~\ref{lem_ttm_perturbation} and~\ref{lem_ttm_kmeans} as follows.
\\{\bf Lemma$~\ref{lem_ttm_perturbation}^*$.}
If $\delta > 2\gamma_n$ for all large $n$, then with probability $\left(1-o(1)\right)$,
\begin{equation}
\left\Vert \overline{X} - ZQ\right\Vert_F \leq \frac{2\gamma_n\sqrt{2kn_1}}{\delta}\;.
\end{equation}
{\bf Lemma$~\ref{lem_ttm_kmeans}^*$.}
If $\delta \geq 2\gamma_n\sqrt{\frac{8kn_1\ln n}{n_k}}$,
then with probability $1-o(1)$,
\begin{equation}
\err = O\left(\frac{kn_1\gamma_n^2}{\delta^2}\right)\;.
\end{equation}
Here, the stronger condition is required for the $k$-means error bound.
Now, observe that we can bound $\gamma_n^2$ as
\begin{align*}
\gamma_n^2 \leq 288\left( \frac{\ln n}{d} + 
\frac{\ln n}{N}\left(1+\frac{2\beta}{d}\right)\right).
\end{align*} 
The above bound immediately implies the error bound in~\eqref{eq_ttm_error_bound_sampled}, whereas the condition in Lemma$~\ref{lem_ttm_kmeans}^*$ is satisfied if~\eqref{eq_dn_condns_sampled} holds.
Thus, the claim of Theorem~\ref{thm_ttm_consistency_sampled} follows.

\subsection{TTM with Iterative Sampling}

In this section, we discuss practical strategies for sampling. The purpose of the section is two-fold. We first relate the weighted sampling strategy~\eqref{eq_ttm_wted_sampling} to heuristics used in practice.
We then suggest a practical variant of Algorithm~\ref{alg_sampledttm} to solve the problem of subspace clustering. Our experimental results in next section will validate the efficacy of this method in realistic problems.

We recall that Corollary~\ref{cor_ttm_sampling_examples} led to the conclusion that, in general, weighted sampling~\eqref{eq_ttm_wted_sampling} achieves a runtime that is smaller than that of uniform sampling by a factor of $\alpha_n$. It is obvious that specifying this distribution involves computing
all edge weights, which in turn, requires a single pass over the adjacency tensor. Even the weighted sampling of~\citet{Bhojanapalli_2015_arxiv_05023} suffers from the same issue.
However, even this is not acceptable in practice as a single pass also has computational complexity of $O(n^m)$, and hence, \ref{alg_sampledttm} based on~\eqref{eq_ttm_wted_sampling} is mainly of theoretical interest.
But our analysis leads to an important conclusion---\emph{sample edges with larger weights more frequently}.

This is essentially the idea commonly used in most tensor based algorithms.
In the case of matching algorithms, one uses an efficient nearest neighbour
search to sample the larger tensor entries~\citep{Duchenne_2011_jour_TPAMI}.
On the other hand, the subspace clustering literature has acknowledged the idea
of iterative sampling~\citep{Chen_2009_jour_JourCV,Jain_2013_conf_ICCV}, 
where one uses an alternating strategy of finding clusters using a sampled set of edges,
and then re-sampling edges for which  at least $(m-1)$ nodes belong to a cluster. 
It is not hard to realise that both sampling techniques give higher preference to edges with large weights,
and hence, as a consequence of Corollary~\ref{cor_ttm_sampling_examples},
both methods are expected to perform better than uniform sampling.
Thus Corollary~\ref{cor_ttm_sampling_examples} provides a theoretical justification
for why such heuristics work, thereby answering an open question posed 
by~\citet{Chen_2009_jour_JourCV}.

We now turn to the problem of designing a practical variant of TTM based on the above discussion. We use the conclusions of Corollary~\ref{cor_ttm_sampling_examples} 
and present an iterative version of Algorithm~\ref{alg_ttm}
for the purpose of subspace clustering. We henceforth refer to this algorithm as \emph{tensor trace maximisation with iterative sampling}, or simply Tetris.
An additional reason for presenting this algorithm is to address a paradoxical situation that arose in our previous work~\citep{Ghoshdastidar_2015_conf_ICML}.
While the theoretical results suggest that TTM perform better than HOSVD based techniques~\citep{Govindu_2005_conf_CVPR,Chen_2009_jour_JourCV},
experiments on large benchmark problems did not align with the same conclusion. We later realised that this disparity occurred because we had combined TTM with a naive sampling technique, but had compared with the practical iterative variant of HOSVD. The numerical comparisons in this paper using Tetris resolves this issue, and shows that TTM is indeed more favourable.

We now present Tetris for solving the subspace clustering problem~\citep{Soltanolkotabi_2014_jour_AnnStat}.
In this problem, 
one is given a collection of $n$ points 
$Y_1,Y_2,\ldots, Y_n\in\R^{r_a}$ in an high dimensional \emph{ambient} space.
However, there are $k$ subspaces, each of dimension at most $r<r_a$, such that one can
represent $Y_i$  as 
\begin{equation*}
Y_i = \widetilde{Y}_i + \eta_i \;,
\end{equation*}
where $\widetilde{Y}_i$ lies in one of the $k$ subspaces, and
$\eta_i$ is a noise term.
The objective of a subspace clustering algorithm is to group $Y_1,\ldots, Y_n$ into $k$
disjoint clusters such that each cluster corresponds to exactly one of the $k$ 
low-dimensional subspaces.
A hypergraph or tensor based subspace clustering approach~\citep{Agarwal_2005_conf_CVPR,Govindu_2005_conf_CVPR}
involves construction of a weighted $m$-uniform hypergraph such that $m\geq(r+2)$
and the weight of an edge $e=\{i_1,\ldots, i_m\}$ is given by
\begin{equation}
w_e = w(\{i_1,\ldots,i_m\}) = \exp\left(-\frac{f_r(Y_{i_1},\ldots, Y_{i_m})}{\sigma^2}\right)\;.
\label{eq_subspace_edgeweight}
\end{equation}
Here, $f_r(\cdot)$ computes the error of fitting a $r$-dimensional subspace
for the given $m$ points, and $\sigma$ is a scaling parameter.
Different choices for $f_r(\cdot)$ has been considered in the literature 
based on Euclidean distance of points from the 
estimated subspace~\citep{Govindu_2005_conf_CVPR,Jain_2013_conf_ICCV},
polar curvature of the points~\citep{Chen_2009_jour_JourCV} among others.
\citet{Chen_2009_jour_JourCV} also proposed a heuristic for estimating $\sigma$
at each iteration.

We present Algorithm~\ref{alg_Tetris} for the subspace clustering problem.
We fix the order of the tensor as $m=(r+2)$, and define $f_r(\cdot)$ in terms of
polar curvature~\citep[see Equations 1-3 of][]{Chen_2009_jour_JourCV}. 
We also incorporate the convergence criteria and the
estimation procedure for $\sigma$ used by~\citet{Chen_2009_jour_JourCV}, 
which are not explicitly stated below.
Furthermore, to standardise with their approach, Tetris
uses a one-sided degree normalisation and computes left singular vectors of the normalised adjacency matrix.

\begin{varalgorithm}{Tetris}
\caption {: TTM with iterative sampling for subspace clustering}
\label{alg_Tetris}
\begin{algorithmic}[1]
 \REQUIRE Dataset $Y = [Y_1,\ldots, Y_n]$; $k=$ Number of subspaces;
 \newline $r=$ Maximum subspace dimension; and
 \newline $c=$ A hyper-parameter controlling number of sampled edges ($N=nc$) \\
 \STATE Set $m=r+2$.
 \STATE Uniformly sample $c$ subsets of $Y$, each  containing $(m-1)$ points.
 \STATE Initialise $\widehat{A}\in\R^{n\times n}$ to a zero matrix.
 \FOR{$j=1$ to $c$}
 \STATE Consider $j^{th}$ subset of $Y$ with the points $Y_{j_1},\ldots, Y_{j_{m-1}}$.
 \FOR{$i=1$ to $n$}
 \STATE Compute the weight $w_e$ for the edge $e=\{Y_i, Y_{j_1},\ldots, Y_{j_{m-1}}\}$
 using \eqref{eq_subspace_edgeweight}.
 \STATE Update $\widehat{A}_{ij_l} = \widehat{A}_{ij_l} + w_e$ for all $l=1,\ldots,m-1$. 
 \ENDFOR
 \ENDFOR
 \STATE Let $\widehat{D}\in\R^{n\times n}$ be diagonal with 
 $\widehat{D}_{ii} = \sum\limits_{j=1}^n \widehat{A}_{ij}$,
   and $\widehat{L} = \widehat{D}^{-1}\widehat{A}$.
 \STATE Compute $k$ dominant left singular vectors of $\widehat{L}$, denoted by 
 $\widehat{X}\in\R^{n\times k}$ .
 \STATE Normalise rows of $\widehat{X}$ to have unit norm.\\
 \STATE Run $k$-means on the rows of the normalised matrix, 
 and partition $Y$ into $k$ clusters.\\
 \STATE From each obtained cluster, sample $c/k$ subsets, each of size $(m-1)$.
 \STATE Repeat from Step 3, and iterate until convergence.
 \ENSURE Clustering of $Y$ into $k$ disjoint clusters.\\
\end{algorithmic}
\end{varalgorithm}

\section{Experimental Validation}
\label{sec_experiment}
In this section, we present numerical illustrations related to uniform hypergraph partitioning.
The numerical results are categorised into four parts.
The first set of results is based on the setting of Corollary~\ref{cor_ttm_consistency_balanced_dense},
and validates our theoretical observations about TTM.
We next compare the performance of TTM with several uniform hypergraph partitioning 
methods for some small scale problems. 
The reason for restricting our study to small problems is because, in such cases, the
hypergraphs can be completely specified, and edge sampling can be avoided.
The rest of the section focuses on practical versions of the subspace 
clustering problem, where we compare sampled variants of TTM with state of the art subspace
clustering algorithms.
Our experiments include both synthetic subspace clustering problems~\citep{Park_2014_conf_NIPS}
and benchmark motion segmentation problem~\citep{Tron_2007_conf_CVPR}.

\subsection{Comparison of Spectral Algorithms}
\label{sec_spec_comp}

We first compare the performance of TTM
with the HOSVD based algorithm~\citep{Govindu_2005_conf_CVPR,Ghoshdastidar_2014_conf_NIPS} 
and the NH-Cut algorithm~\citep{Zhou_2007_conf_NIPS,Ghoshdastidar_2017_jour_AnnStat}.
This study is based on the model related to Corollary~\ref{cor_ttm_consistency_balanced_dense},
where a $m$-uniform hypergraph is generated on $n$ vertices.
We assume here that $\alpha_n=1$, $k=2$, and the true clusters are of equal size.
The edges occur with following probabilities. 
If all vertices in an edge do not belong to the same cluster, then the edge probability is
$q=0.2$, else it is $(p+q)$ for some $p\in(0,1-q)$ specified below.

\begin{figure}[t]
\centering
 \begin{tabular}{cc}
 \multicolumn{2}{l}{~~~~~~~~~~~~~(a) $m=2$~~~~~~~~~~~~~~~~~~~(b) $m=3$~~~~~~~~~~~~~~~~~~(c) $m=4$} \\
 \rotatebox{90}{\hspace{5mm} Error, $\err$} &
 \includegraphics[clip=true, trim=40mm 0mm 65mm 0mm, width=0.9\textwidth]{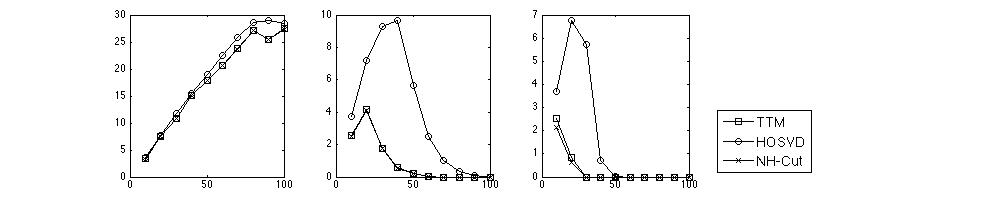}\\
 \multicolumn{2}{c}{Total number of vertices, $n$} \\
 \end{tabular}
 \caption{Number of vertices mis-clustered by TTM, HOSVD and NH-Cut as $n$ increases. 
 The figures from left to right correspond to cases with $m=2$, 3 and 4, respectively.}
  \label{fig_vary_m}
\end{figure}

In Figure~\ref{fig_vary_m}, we show results for three examples, where $p$ is fixed at $p=0.1$,
$m$ is varied over $m=2,3,4$, and the total number of vertices $n$ grows from 10 to 100.
For each case, 50 planted hypergraphs are generated, and subsequently partitioned 
by TTM, HOSVD and NH-Cut. 
The mean error, $\err$, is reported for each algorithm as a function of $n$.
Figure~\ref{fig_vary_m} shows that the performance of TTM and NH-Cut are similar,
and the errors incurred by these methods are significantly smaller than that of HOSVD. 
This observation validates Corollary~\ref{cor_ttm_consistency_balanced_dense} and Remark~\ref{rem_spectral_comp}.
It can also be seen empirically that all three methods have a sub-linear error rate for $m=2$,
\ie they are weakly consistent, whereas, $\err=o(1)$ for $m\geq3$.

\begin{figure}[t]
\centering
 \begin{tabular}{cc}
 \multicolumn{2}{l}{~~~~~~~~~~~~~(a) $p=0.1$~~~~~~~~~~~~~~~~(b) $p=0.05$~~~~~~~~~~~~~~~(c) $p=0.025$} \\
 \rotatebox{90}{\hspace{5mm} Error, $\err$} &
 \includegraphics[clip=true, trim=41mm 0mm 64mm 0mm, width=0.9\textwidth]{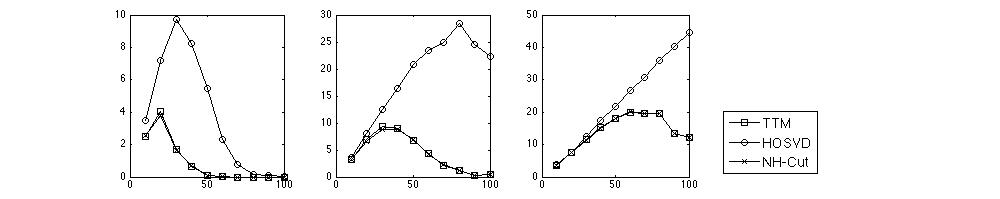}\\
 \multicolumn{2}{c}{Total number of vertices, $n$} \\
 \end{tabular}
 \caption{Number of vertices mis-clustered by TTM, HOSVD and NH-Cut as $n$ increases. 
 The figures from left to right correspond to cases with $p=0.1$, 0.05 and 0.025, respectively.}
  \label{fig_vary_p}
\end{figure}

We consider another example on bi-partitioning 3-uniform hypergraphs, where we fix $q=0.2$ but the gap $p$ is decreased as 0.1, 0.05 and 0.025. 
Figure~\ref{fig_vary_p} shows the errors, averaged over 50 runs, incurred by the three methods
as the hypergraph grows.
Note that the problem becomes harder as $p$ reduces, and the  performance of HOSVD is 
highly affected. But, the effect is much less in case of TTM and NH-Cut.
This follows from Theorem~\ref{thm_ttm_consistency}, where one can observe that,
in the present context $\err$ varies as $1/p^2$. Same holds for NH-Cut, but in the
case of HOSVD, $\err$ varies as $1/p^4$ making the algorithm more sensitive to reduction
in probability gap.

\subsection{Comparison of Hypergraph Partitioning Methods}

We consider similar studies with other hypergraph partitioning methods, such as
methods based on symmetric non-negative tensor factorisation (SNTF)~\citep{Shashua_2006_conf_ECCV},
higher order game theoretic clustering (HGT)~\citep{RotaBulo_2013_jour_TPAMI}
and the hMETIS algorithm widely used in VLSI community~\citep{Karypis_2000_jour_VLSI}.
For the latter two methods, we have used implementations provided by the authors.

\begin{figure}[ht]
\centering
 \begin{tabular}{cc}
 \rotatebox{90}{\small Probability gap $p$} &
 \includegraphics[clip=true, trim=36mm 10mm 15mm 5mm, width=0.92\textwidth]{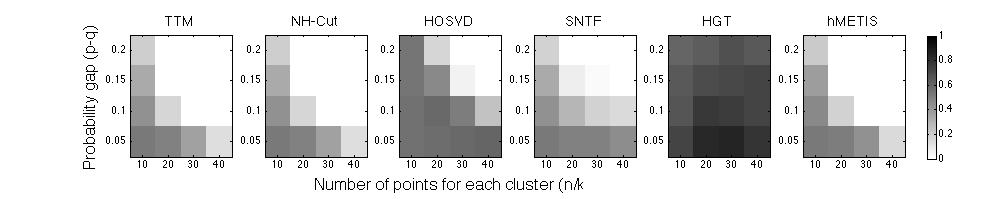}\\
 & Number of vertices in each cluster, $n/k$ \\
 \end{tabular}
 \caption{Fractional error incurred by hypergraph partitioning algorithms under a planted model.
 The cluster size, $(n/k)$, and the probability gap $p$ are varied. 
 The colour bar indicates the shade corresponding to different levels of error, with darker shade
 representing larger error.}
 \label{fig_hyp_planted}
\end{figure}

We first compare the different algorithms under a planted model for 3-uniform hypergraphs
with $k=3$ planted clusters of equal size.
As before, we assume the hypergraph to be dense, $\alpha_n=1$, and the inter-cluster edges
occur with probability $q=0.2$. 
We study the performance of the methods as the number of vertices $n$, and the 
probability gap $p$ varies. 
The fractional clustering error, $\frac1n\err$, averaged over 50 runs, is reported in
Figure~\ref{fig_hyp_planted}.

The figure shows the previously observed trends for TTM, NH-Cut and HOSVD.
In addition, it is observed that SNTF and hMETIS provide nearly similar,
but marginally worse results than TTM.
However, HGT uses a greedy strategy for extracting individual clusters,
and hence, often identifies a majority of the vertices as outliers,
thereby resulting in poor performance.

\begin{figure}[t]
\centering
 \begin{tabular}{cc}
 \rotatebox{90}{\small Noise level, $\sigma_a$} &
 \includegraphics[clip=true, trim=36mm 10mm 15mm 5mm, width=0.92\textwidth]{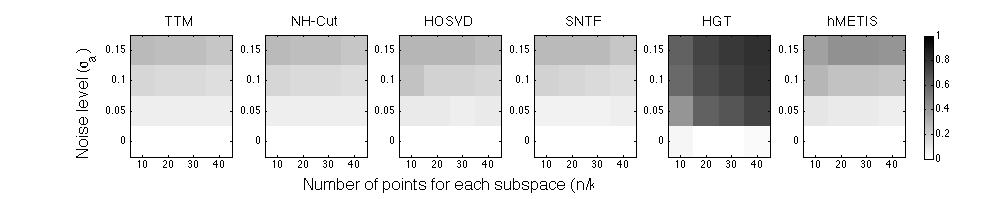}\\
 & Number of points in each subspace, $n/k$ \\
 \end{tabular}
 \caption{Fractional error incurred by hypergraph partitioning algorithms in clustering 
 noisy points from three intersecting lines.
 The cluster size, $(n/k)$, and the noise level $\sigma_a$ are varied. 
 The colour bar indicates the shade corresponding to different levels of error, with darker shade
 representing larger error.}
 \label{fig_hyp_subspace}
\end{figure}

Since partitioning algorithms find use in a variety of applications,
we compare the performance of the above algorithms for the subspace clustering problem.
In particular, we consider the line clustering problem in an ambient space of
dimension 3.  
We randomly generate three one-dimensional subspaces, and sampled $n/k$
random points from each subspace.
As mentioned in the previous section, the data points in subspace clustering problems 
are typically perturbed by noise.
To simulate this behaviour, we add a zero mean Gaussian noise vector to each point.
The covariance of the noise vectors is given as $\sigma_aI$, where we vary
$\sigma_a$ to control the difficulty of the problem.
We construct a weighted 3-uniform similarity hypergraph based on 
polar curvature of triplet of points, which is partitioned by the different methods.
The fractional clustering errors are presented in Figure~\ref{fig_hyp_subspace}.
As expected, all the methods can identify the exact subspace in the absence of
noise, and the errors increase for larger $\sigma_a$.
Apart from HGT, a good performance is observed from all the methods.

One can observe that the above comparisons were based on very small problems,
where the hypergraph consists of at most 120 vertices.
This restriction was imposed since specification of the entire affinity tensor
is computationally infeasible for large hypergraphs.
To demonstrate the performance of TTM in practical settings, we study its 
sampled variants in the subsequent sections.

\subsection{Comparison of Subspace Clustering Algorithms: Synthetic Data}

We now compare our method against the state of the art subspace clustering algorithms. 
We consider sampled variants of the TTM algorithm, \ie TTM with uniform sampling
and TTM with iterative sampling (Tetris). 
We note that from practical consideration, we do not consider
aforementioned hypergraph partitioning methods that require computation of the entire tensor.
The clustering algorithms under consideration include:
\begin{itemize}
\item $k$-means algorithm for clustering based on Euclidean distance,
\item $k$-flats algorithm~\citep{Bradley_2000_jour_JourGlboalOpt} which generalises $k$-means to subspace clustering, 
\item sparse subspace clustering (SSC)~\citep{Elhamifar_2013_jour_TPAMI}, which finds
clusters by estimating the subspaces,
\item subspace clustering using low-rank representation (LRR)~\citep{Liu_2010_conf_ICML}, 
\item thresholding based subspace clustering (TSC)~\citep{Heckel_2013_conf_ICASSP},
\item faster variant of SSC using orthogonal matching pursuit (SSC-OMP)~\citep{Dyer_2013_jour_JMLR},
\item greedy subspace clustering using nearest subspace neighbour search and spectral clustering (NSN+Spectral)~\citep{Park_2014_conf_NIPS},
\item spectral curvature clustering (SCC)~\citep{Chen_2009_jour_JourCV}, which is
an iterative variant of HOSVD,
\item sparse Grassmann clustering (SGC)~\citep{Jain_2013_conf_ICCV},\footnote{
For this method, we have used our implementation.} yet another
variation of HOSVD where some information about the eigenvectors computed
in previous iterations is retained,
\item Algorithm~\ref{alg_Tetris}, and
\item Algorithm~\ref{alg_ttm} with uniform sampling, which is derived by performing
a single iteration of
Steps 1-14 of Algorithm~\ref{alg_Tetris}.\footnote{Here, the edge sampling is not exactly 
uniform since we only select the $c$ subsets of size $(m-1)$ uniformly.}
\end{itemize}

\begin{figure}[t]
\centering
 \begin{tabular}{cc}
 \rotatebox{90}{\hspace{15mm} Noise level, $\sigma_a$} &
 \includegraphics[clip=true, trim=36mm 10mm 15mm 5mm, width=0.9\textwidth]{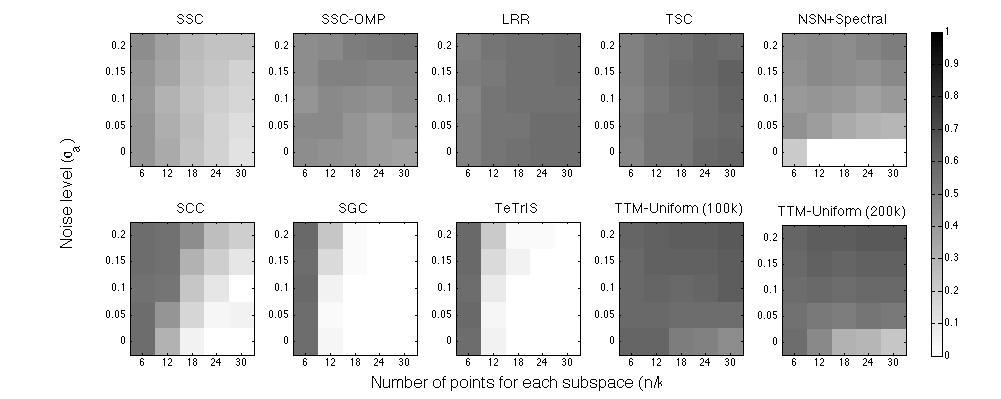}\\
 & Number of points in each subspace, $n/k$ \\
 \end{tabular}
 \caption{Fractional error incurred by subspace clustering algorithms for synthetic data.
 The number of points in each subspace, $(n/k)$, and the variance of the noise vector 
 $\sigma_a$ is varied. The colour bar indicates the shades for different levels of error.}
 \label{fig_expt_synthetic}
\end{figure}

We first focus on the problem of clustering randomly generated subspaces.\footnote{The experimental setup has been adapted from~\citet{Park_2014_conf_NIPS},
and the codes are available at: \\
\url{http://sml.csa.iisc.ernet.in/SML/code/Feb16TensorTraceMax.zip}}
In an ambient space of dimension $r_a=5$, 
we randomly generate $k=5$ subspaces each of dimension $r=3$. From each subspace,
we randomly sample $n/k$ points and perturb every point with a 
5-dimensional Gaussian noise vector with mean zero and covariance $\sigma_aI$.
In Figure~\ref{fig_expt_synthetic}, we report the fractional error, $\frac1n\err$, incurred by 
various subspace clustering algorithms when $(n/k)$ and $\sigma_a$ are varied.
The results are averaged over 50 independent trials.
We note that for existing methods, we fix the parameters as mentioned in~\citet{Park_2014_conf_NIPS}.
For Tetris and SGC, the parameters are set to the same values as SCC,
where $c=100k$ and $\sigma$ as in~\eqref{eq_subspace_edgeweight} is determined by 
the algorithm. In case of uniformly sampled TTM, we fix $\sigma$ 
to be same as the value determined by Tetris. 
To demonstrate that sampling more edges lead to error reduction,
we consider uniform sampling for two values $c = 100k$ and $200k$. 

Figure~\ref{fig_expt_synthetic} shows that Tetris and SGC clearly outperform other methods
over a wide range of settings. In particular, it can be seen that greedy methods like NSN
is accurate in the absence of noise, but a drastic increase in error occurs when the data is 
noisy. The effect of noise is much less in hypergraph based methods like SCC, SGC or Tetris.
One can also observe that the hypergraph based methods do not work well when there are very few 
points in each cluster (for example, 6). This is expected since, by definition, these algorithms 
construct 5-uniform hypergraphs $(m=r+2)$ in this case, and hence, there are very few
edges $(\binom65 = 6)$ with large weight for each cluster. However, with increase in 
number of points, there is a rapid decay in the clustering error.
This also shows the consistency of these methods empirically. To this end, it seems that NSN or SSC
should be recommended for small scale problems (smaller $n/k$), 
whereas Tetris or SGC should be the algorithm of choice for larger $n$
and possible presence of noise.
Finally, we also observe that TTM with uniform sampling, even with twice the
number of samples, performs quite poorly as compared to Tetris.
However, with increase in the number of sampled edges, some extent of error reduction is observed.

\subsection{Comparison of Subspace Clustering Algorithms: Motion Segmentation}

The Hopkins 155 database~\citep{Tron_2007_conf_CVPR} contains a number of videos
capturing motion of multiple objects or rigid bodies. In each video, few features are
tracked along the frames, each giving rise to a motion trajectory that resides in a space
of dimension twice the number of frames. One can show 
that under particular camera models, all trajectories corresponding to a particular rigid body
motion span a subspace of dimension at most four~\citep{Tomasi_1992_jour_JourCV}.
Thus, the problem of segmenting different motions in a video can be posed as a subspace clustering problem.

The Hopkins database contains 120 sequences, each containing two motions, and 35
three motion sequences. We run above mentioned subspace clustering algorithms
for purpose of motion segmentation. For existing approaches, the parameters
specified in~\citet{Park_2014_conf_NIPS} have been used, and for Tetris and SGC, we use the parameters for SCC. TTM with uniform sampling is not considered due to 
its higher error rate. 
Table~\ref{tab_expt_motion} reports the mean and median of the percentage errors incurred by 
different algorithms, where these statistics are computed over all 2-motion and 3-motion
sequences. 
In order to remove the effect of randomisation due to sampling (for SCC, SGC, Tetris)
or initialisation (for $k$-means, $k$-flats, NSN), we average the results over 20 independent trials. 
The average computational time (in seconds) of each algorithm for each video is also reported.\footnote{%
We note that the reported time is based on the fact that we have used Matlab
implementations of the algorithms, run on a Mac OS X operating system with 2.2 GHz Intel 
Core i7 processor and 16 GB memory.}

Table~\ref{tab_expt_motion} shows that Tetris performs quite well in comparison 
with state of the art subspace clustering algorithms.
In particular, Tetris achieves least mean error for the two motion problem.
The computational time for Tetris is also much smaller than other accurate methods
like SSC and LRR. The mean error achieved by Tetris is also smaller than SCC in either cases.
We note here that the best known 
results for Hopkins 155 database is achieved by the algorithm in~\citet{Jung_2014_conf_CVPR}, which 
uses techniques based on epi-polar geometry, and hence, it is not a subspace clustering algorithm. 
Smaller errors have also been reported in the literature when one construct larger tensors,
$m=8$~\citep{Jain_2013_conf_ICCV}, or
uses manual tuning of hyper-parameters~\citep{Ghoshdastidar_2015_conf_AAAI}.
However, in either cases, computational time increases considerably.

\begin{table}[t]
\centering
 \begin{tabular}{|c||c|c|c||c|c|c|}
 \hline
 Algorithm & \multicolumn{3}{c||}{2 motion (120 sequences)} 
 & \multicolumn{3}{c|}{3 motion (35 sequences)}\\
 \cline{2-7}
 & Mean (\%) & Median (\%) & Time (s) & Mean (\%) & Median  (\%) & Time (s) \\
 \hline\hline
 $k$-means	& 19.58	& 17.92	& 0.03	& 26.13	& 20.48	& 0.05 \\
 $k$-flats		& 13.19 	& 10.01	& 0.38	& 15.45	& 14.88	& 0.76 \\
 SSC			& 1.53	& 0.00	& 0.80	& 4.40	& 0.56	& 1.51 \\
 LRR			& 2.13	& 0.00	& 0.94	& 4.03	& 1.43	& 1.29 \\
 SSC-OMP	& 16.93	& 13.28	& 0.72	& 27.61	& 23.79	& 1.23 \\
 TSC			& 18.44	& 16.92	& 0.19	& 28.58	& 29.67	& 0.51 \\
 NSN+Spec	& 3.62	& 0.00	& 0.08	& 8.28	& 2.76	& 0.17 \\
 SCC		& 2.53	& 0.03	& 0.45	& 6.40	& 1.46	& 0.76 \\
 SGC		& 3.50	& 0.41	& 0.54	& 9.08	& 5.05	& 0.89 \\
 Tetris 		& 1.31	& 0.02	& 0.50	& 5.71	& 1.19	& 0.90 \\
 \hline
 \end{tabular}
 \caption{Mean and median of clustering error and computational time for different subspace 
 clustering algorithms on Hopkins 155 database.}
 \label{tab_expt_motion}
\end{table}

\section{Conclusion}
\label{sec_conclusion}

In this paper, we studied the problem of partitioning uniform hypergraphs that arises
in several applications in computer vision and databases.
We formalised the problem by defining a normalised associativity of a partition in a uniform hypergraph that extends a 
similar notion used in graph literature. 
We showed that the task of finding a partition that maximises normalised associativity
is equivalent to a tensor trace maximisation problem.
We proposed a tensor spectral algorithm (TTM) to solve this problem,
and following the lines of our previous works,
we showed that the TTM algorithm is consistent under a planted partition model.
To this end, we extended the existing model of~\citet{Ghoshdastidar_2014_conf_NIPS,Ghoshdastidar_2017_jour_AnnStat}
by allowing both sparsity of edges as well as the possibility of weighted edges.
Accounting for these factors makes the model more appropriate in the context of computer vision applications.
We derived error bounds for the TTM approach under the planted partition model.
Our bounds indicate that under mild assumptions on the sparsity of the hypergraph, 
TTM is weakly consistent, and
the error bound for TTM is comparable that of NH-Cut, but better than the error rates for HOSVD. This fact is also validated numerically.

Weighted uniform hypergraphs have been particularly interesting in the
computer vision since hypergraphs provide a natural way to represent multi-way similarities.
Yet, it is computationally expensive to compute the entire affinity tensor
of the hypergraph.
As a consequence, several tensor sampling strategies have come into existence.
We provide the first theoretical analysis of such sampling techniques in the context of uniform hypergraph partitioning.
Our result suggests that consistency can be achieved even with very few sampled edges provided that one assigns a higher sampling probability for edges  with larger weight.
The derived sampling rate is much lower than that known in tensor literature~\citep{Bhojanapalli_2015_arxiv_05023,Jain_2014_conf_NIPS}, and our analysis also justifies the superior performance of popular sampling heuristics~\citep{Chen_2009_jour_JourCV,Duchenne_2011_jour_TPAMI}.
We finally proposed a iteratively sampled variant of TTM, and empirically demonstrated the
potential of this method in subspace clustering and motion segmentation applications.

We conclude with the remark that this paper was motivated by practical aspects of hypergraph partitioning, and our aim was to present an approach that can be analysed theoretically, and at the same time, can compete with practical methods. While the paper manages to achieve this goal, there are several questions that still remain unanswered. We list some of the important open problems:

\vskip1ex\noindent {\it(i)} 
What is the threshold for detecting planted partitions in hypergraphs?
\\
This question was recently answered for a special model in the case of bi-partitioning~\citep{Florescu_2016_conf_COLT}, but more general cases have still not been explored.  
For instance, it would interesting to know any extension of the results in~\citet{Abbe_2016_conf_NIPS} for unweighted hypergraphs.
However, this would still not provide any information in the weighted case.
We noted that one of the reasons for our condition in~\eqref{eq_dn_condns}
being worse by logarithmic factors was our analysis is for weighted hypergraphs. It is not known yet whether one can do better for weighted hypergraphs without making distributional assumptions on the edge weights.

\vskip1ex\noindent {\it(ii)} 
What are the complicated examples? How can we deal with these cases? 
\\
Our error bound leads to a counter-intuitive conclusion that even after collapsing the tensor to a matrix, one can get better performance than direct tensor decomposition (HOSVD). The caveat here is that this result is stated for a special case, and still leaves the possibility that that there are models which do not satisfy the condition $\delta>0$, and hence, TTM does not work. 
In~\citet{Ghoshdastidar_2017_jour_AnnStat}, we constructed few examples, but it is not known whether HOSVD works in these cases. We also mention that the use of other tensor methods (such as power iterations) have not been analysed yet. While we showed that existing results cannot be applied due to the difference in assumptions, we do not claim that alternative tensor methods are inapplicable.

On a similar note, it is known in the case of graph partitioning that the eigen gap (quantified by $\delta$ in our results) is not the correct quantity that reflects the presence of a partition.
For instance, there are planted graph models, which clearly reflect a separation of the classes, and yet $\delta$ is not positive. It would be interesting to see how one can deal with such examples in the case of hypergraphs. An alternative approach in such cases is to quantify the separation by the minimum difference (or distance) between the $k$ rows of $\Bb$, which is a matrix for graphs.
For uniform hypergraphs, a similar quantity would be the difference between the $(m-1)$-order slices of the tensor $\Bb$, but it is not clear yet how such a condition can be incorporated into the analysis of hypergraph partitioning.

\vskip1ex\noindent {\it(iii)} 
What is the minimum sample size for achieving weak consistency?
\\
Similar in spirit to the first problem, one can ask what is the minimum sampling rate required to achieve weak consistency irrespective of the partitioning algorithm. We have not made any attempt to answer this question,
and surprisingly, the question is open even in graph partitioning. While column sampling techniques~\citep{Drineas_2006_jour_SICOMP} are often used in conjunction with spectral clustering, the error rate for this combination is not known.

\vskip1ex\noindent {\it(iv)} 
How well does TTM extend for other higher order learning problems?
\\
While TTM unifies a wide variety of higher order learning methods, the special cases studies in this paper were restricted to clustering problems.
However, other problems like tensor matching can also be modelled as partitioning uniform hypergraph with two partitions of widely different sizes~\citep{Duchenne_2011_jour_TPAMI}. 
In principle, this problem has a flavour quite similar to that of the 
well known planted clique problem~\citep{Alon_1998_conf_SODA}.
The state of the art algorithms in tensor matching rely on tensor power iterations, and in~\citet{Ghoshdastidar_2017_jour_AnnStat}, we showed that a naive use of NH-Cut does not fare well. 
We feel that in this particular setting, better theoretical guarantees can be achieved by power iteration based approaches.

\acks{%
We thank the reviewers for useful suggestions and key references.
During this work, D. Ghoshdastidar was supported by the 2013 Google Ph.D. Fellowship in Statistical Learning Theory. 
This work is also partly supported by the Science and Engineering Research Board through the grant SERB:SB/S3/EECE/093/2014.}

\appendix
\section{Proofs of Technical Lemmas and Corollaries}

Here, we sequentially  present the proofs of lemmas and corollaries stated in the paper.
We begin with the calculations that lead to~\eqref{eq_partition_assoc_tensor}.

\subsection{Proof of Equation~\eqref{eq_partition_assoc_tensor}}
We recall that the entries of $Y^{(l)}$ is simply $Y_{ij}^{(l)} =  \frac{\1\{i\in\Vc_j\}}{(\vol(\Vc_j))^{\beta_l}}$.
The claimed relation follows by computing diagonal entries of
$\Ab \times_1 {Y^{(1)}}^T \times_2 {Y^{(2)}}^T
 \times_3 \ldots \times_m {Y^{(m)}}^T$, which are given by
\begin{align*}
 \left(\Ab \times_1 {Y^{(1)}}^T \times_2 \ldots \times_m {Y^{(m)}}^T\right)_{j\ldots j}
 &= \sum_{i_1,\ldots,i_m=1}^n \Ab_{i_1\ldots i_m} Y_{i_1j}^{(1)} \ldots Y_{i_m j}^{(m)}
 \\& = m! \sum_{i_1<\ldots<i_m} \Ab_{i_1\ldots i_m} \frac{\1\{i_1,\ldots,i_m\in \Vc_j\}}{(\vol(\Vc_j))^{\beta_1+\ldots+\beta_m}} \;,
\end{align*}
where we use the symmetry of the terms to group all $m!$ permutations of every distinct $i_1,\ldots, i_m$.
Since $\sum_l \beta_l =1$, the denominator is simply $\vol(\Vc_j)$,
whereas the indicator in the numerator counts only the edges $e\subset\Vc_j$
and $\Ab_{i_1\ldots i_m}$ provides the weights. So the above quantity is simply $m!\frac{\assoc(\Vc_j)}{\vol(\Vc_j)}$, and summing over all diagonal entries results in the normalised associativity, scaled by $m!$.

\subsection{Proof of Corollary~\ref{cor_ttm_consistency_balanced_dense}}
We begin by computing $\Ac$ as defined in~\eqref{eq_Ac_expn}
\begin{align*}
 \Ac_{ij} = \left\{\begin{array}{ll}
 (m-2)!\alpha_n \left(p\binom{\frac{n}{k}-2}{m-2} + q\binom{n-2}{m-2}\right)
 & \text{if } i \neq j, \psi_i = \psi_j
 \\\\
 (m-2)!\alpha_n q\binom{n-2}{m-2}
 & \text{if } i \neq j, \psi_i \neq \psi_j
 \\\\
 0
 & \text{if } i = j.
                 \end{array}\right.
\end{align*}
From the definition of $G$, $\Dmin$ and $\delta$, one can compute that
\begin{align}
 \Dmin = (m-1)!\alpha_n \left(p\binom{\frac{n}{k}-1}{m-1} + q\binom{n-1}{m-1}\right) \;,
\label{eq_example_Dmin}
\end{align}
and
\begin{align}
\delta= \lambda_k(G) \frac{n}{k\Dmin} = \frac{(m-2)!\alpha_n p n}{k\Dmin}\binom{\frac{n}{k}-2}{m-2}
\label{eq_example_delta}
\end{align}
We need to validate that the conditions in~\eqref{eq_dn_condns}, 
or equivalently~\eqref{eq_Dmin_condns}, are satisfied.
Given $\alpha_n=1$, one can see that $\Dmin = \Theta(n^{m-1})$
 easily satisfies first condition of~\eqref{eq_Dmin_condns} for large $n$.\footnote{%
The notation $f_n=\Theta(n)$ denotes that there exists constants $c, C$
such that $c n \leq f_n \leq Cn$ for all large $n$.} 
 Also
\begin{align*}
 \delta^2 \Dmin = \Theta\left( \left(\frac{n}{k}\right)^{2m-2}\frac1\Dmin \right) = 
 \Theta\left( \frac{n^{m-1}}{k^{2m-2}} \right) 
 = \Omega\left(n^{(m-1)/2}(\ln n)^{2m-2}\right)\;,
\end{align*}
taking into account that $k=O\left(\frac{n^{1/4}}{\ln n}\right)$.
Thus, the second condition in~\eqref{eq_Dmin_condns} also holds for large $n$ and for all $m\geq2$.
Subsequently, one can applying the bound in~\eqref{eq_ttm_error_bound} to
claim that 
$\err = O\left(\frac{n^{(3-m)/2}}{(\ln n)^{2m-3}}\right)$
with probability $(1-o(1))$. 

\subsection{Proof of Corollary~\ref{cor_ttm_consistency_balanced_sparse}} 
For $k=O(\ln n)$, one can verify that~\eqref{eq_dn_condns} holds if
$\Dmin = \Omega\left( \frac{(\ln n)^3}{\delta^2} \right)$.
From~\eqref{eq_example_Dmin} and~\eqref{eq_example_delta}, we have
$\Dmin = \Theta(\alpha_n n^{m-1})$ and 
$\delta^2\Dmin = \Omega\left(\alpha_n n^{m-1} (\ln n)^{2-2m}\right)$.
Hence, choosing $\alpha_n \geq \frac{C(\ln n)^{2m+1}}{n^{m-1}}$ for sufficiently large $C$
ensures that~\eqref{eq_Dmin_condns} is satisfied.
Subsequently with probability $\left(1 - O\left(n^{-2}+ (\ln n)^{-1/4}\right)\right) = (1-o(1))$,
we obtain an error bound
\begin{align*}
\err = O\left(\frac{n\ln n}{\delta^2\Dmin}\right) = 
O\left(\frac{n\ln n}{\alpha_n n^{m-1} (\ln n)^{2-2m}}\right) = O\left(\frac{n}{(\ln n)^2}\right) =o(n)\;,
\end{align*}
which completes the proof.

\subsection{Proof of Lemma~\ref{lem_ttm_expected}}
The proof is along the lines of the proof of Lemma~4.5 in~\citet{Ghoshdastidar_2017_jour_AnnStat}.
We still include a sketch of the proof since the quantities involved are different from the terms dealt in the mentioned paper.
From the discussions following~\eqref{eq_Ac_expn}, one can see that the matrix
$\Ac$ may be expressed as
\begin{equation*}
 \Ac = ZGZ^T - J\;,
\end{equation*}
where $J\in\R^{n\times n}$ is diagonal with $J_{ii} = G_{\psi_i\psi_i}$.
Following the arguments of~\citet{Ghoshdastidar_2017_jour_AnnStat}, one can show that
there is a matrix $\mathcal{G}\in\R^{k\times k}$ with eigen decomposition 
$\mathcal{G} = U\Lambda_1 U^T$ such that
$\Lc\Xc = \Xc\Lambda_1$,
where $\Xc = Z(Z^TZ)^{-1/2}U$, and it satisfies $\Xc^T\Xc = I$.
Thus, the columns of $\Xc$ are orthonormal eigenvectors of 
$\Lc$ corresponding to the eigenvalues in $\Lambda_1$. 
It is also known that the other $(n-k)$ orthonormal eigenvectors correspond to eigenvalues
from the set $\left\{-\frac{J_{ii}}{\Dc_{ii}}:{1\leq i\leq n}\right\}$.

Thus, to prove the claim we need to ensure that $\Xc$ corresponds to the dominant eigenvectors of $\Lc$, or in other words,
\begin{equation}
 \lambda_k(\mathcal{G}) > \max_{1\leq i \leq n} \left(-\frac{J_{ii}}{\Dc_{ii}}\right)
 =  -\min_{1\leq i \leq n} \frac{G_{\psi_i\psi_i}}{\Dc_{ii}} \;.
 \label{eq_ttm_eigengap_condn}
\end{equation}
A lower bound on $\lambda_k(\mathcal{G})$ can be derived from Rayleigh's principle
and Weyl's inequality as
\begin{equation*}
 \lambda_k(\mathcal{G}) \geq \lambda_k(G) \min_{1\leq i \leq n} \frac{(Z^TZ)_{\psi_i\psi_i}}{\Dc_{ii}}  - \max_{1\leq i \leq n} \frac{G_{\psi_i\psi_i}}{\Dc_{ii}} \;.
\end{equation*}
Noting that $(Z^TZ)_{j j} = n_{j}$, size of cluster-$j$, one can readily see from above that if $\delta>0$,
then~\eqref{eq_ttm_eigengap_condn} is satisfied and hence, $\Xc$ contains the dominant
eigenvectors. It is also useful to note that 
$\delta \leq \lambda_k(\Lc) - \lambda_{k+1}(\Lc)$,
\ie $\delta$ is a lower bound on the eigen gap between the $k^{th}$ and $(k+1)^{th}$ largest
eigenvalues of $\Lc$. This fact is used later in the proof of 
Lemma~\ref{lem_ttm_perturbation}.

\subsection{Proof of Lemma~\ref{lem_ttm_random_deviation}}
We begin the  proof with the claims that if $\Dmin>9(m-1)!\ln n$, then
\begin{equation}
 \P_H \left( \max_{1\leq i\leq n} \left| \frac{D_{ii}}{\Dc_{ii}} - 1\right|
 > 3\sqrt{\frac{(m-1)!\ln n}{\Dmin}}\right) \leq \frac{2}{n^2} \;,
 \label{eq_D_concentration}
\end{equation}
and
\begin{equation}
 \P_H \left( \Vert \Dc^{-1/2}(A-\Ac)\Dc^{-1/2}\Vert_2
 > 3\sqrt{\frac{(m-1)!\ln n}{\Dmin}}\right) \leq \frac{2}{n^2} \;.
 \label{eq_A_concentration}
\end{equation}
We now bound $\Vert L - \Lc \Vert_2$ as
\begin{align*}
 \Vert L - \Lc \Vert_2 
 &\leq \Vert \Dc^{-1/2}A\Dc^{-1/2} - D^{-1/2}AD^{-1/2} \Vert_2 +
   \Vert \Dc^{-1/2}A\Dc^{-1/2} - \Dc^{-1/2}\Ac\Dc^{-1/2} \Vert_2 \;.
\end{align*}
We expand the first term as
\begin{align*}
\Vert \Dc^{-1/2}A\Dc^{-1/2} - &D^{-1/2}AD^{-1/2}\Vert_2
 \\&\leq \Vert (\Dc^{-1/2} - D^{-1/2})A\Dc^{-1/2} + D^{-1/2}A(\Dc^{-1/2} - D^{-1/2})\Vert_2
 \\&\leq \Vert (\Dc^{-1}D)^{1/2} - I\Vert_2\Vert (D\Dc^{-1})^{1/2}\Vert_2 + \Vert(\Dc^{-1}D)^{1/2} - I\Vert_2 ,
\end{align*}
where the last inequality follows since $\Vert D^{-1/2}AD^{-1/2}\Vert_2 = 1$.
Also, note that 
\begin{align*}
 \Vert (\Dc^{-1}D)^{1/2} - I\Vert_2 
 = \max_{1\leq i \leq n} \left| \sqrt{\frac{D_{ii}}{\Dc_{ii}}} - 1\right|
 \leq \max_{1\leq i \leq n} \left| {\frac{D_{ii}}{\Dc_{ii}}} - 1\right|.
\end{align*}
Combining above arguments, we can write
\begin{align}
 \Vert L - \Lc \Vert_2 
 \leq \max_{1\leq i \leq n} \left| {\frac{D_{ii}}{\Dc_{ii}}} - 1\right|
 \left(2+ \max_{1\leq i \leq n} \left| {\frac{D_{ii}}{\Dc_{ii}}} - 1\right|\right)
 + \Vert \Dc^{-1/2}A\Dc^{-1/2} - \Dc^{-1/2}\Ac\Dc^{-1/2} \Vert_2.
 \label{eq_LLc_bound}
\end{align}
Using the bounds in~\eqref{eq_D_concentration} and~\eqref{eq_A_concentration}
along with the fact that $3\sqrt{\frac{(m-1)!\ln n}{\Dmin}} < 1$, one arrives at the claim.

We now prove the concentration bound in~\eqref{eq_D_concentration}. Observe that
\begin{equation*}
 D_{ii} = \sum_{j=1}^n A_{ij} = \sum_{i_2,\ldots,i_m=1}^n \Ab_{ii_2\ldots i_m}
 = (m-1)! \sum_{e\in\Ec: e\ni i} w_e\;,
\end{equation*}
where the last equality holds since the summation over all $i_2,\ldots,i_m$ 
counts each edge containing node-$i$ $(m-1)!$ times.
Since, $D_{ii}$ is a sum of independent random variables, we can use Bernstein inequality
to obtain for any $t>0$,
\begin{align}
 \P_H \left( |D_{ii} - \Dc_{ii}| > t\Dc_{ii} \right)
 &= \P_H \left( \left| \sum_{e\in\Ec: e\ni i} w_e - \E_H [w_e] \right| > \frac{t\Dc_{ii}}{(m-1)!} \right)
 \nonumber
 \\&\leq 2\exp\left(\frac{-\left(\frac{t\Dc_{ii}}{(m-1)!}\right)^2}{
 2\sum\limits_{e\in\Ec: e\ni i} \Var_H(w_e) + \frac{2}{3}\frac{t\Dc_{ii}}{(m-1)!}}\right) \;.
 \label{eq_proof_Dbound1}
\end{align}
Since $w_e \in[0,1]$, we have
\begin{equation*}
 \sum\limits_{e\in\Ec: e\ni i} \Var_H(w_e) \leq \sum\limits_{e\in\Ec: e\ni i} \E_H[w_e] = \frac{\Dc_{ii}}{(m-1)!}\;.
\end{equation*}
Substituting this in~\eqref{eq_proof_Dbound1}, we have
\begin{align*}
 \P_H \left( |D_{ii} - \Dc_{ii}| > t\Dc_{ii} \right) 
 \leq 2\exp\left( - \frac{t^2 \Dc_{ii}}{3(m-1)!}\right) 
 \leq 2\exp\left( - \frac{t^2 \Dmin}{3(m-1)!}\right) .
\end{align*}
The bound in~\eqref{eq_D_concentration} follows from above by setting 
$t = 3\sqrt{\frac{(m-1)!\ln n}{\Dmin}}$, and using a union bound over all $i=1,\ldots,n$. 

Finally, we derive~\eqref{eq_A_concentration} using matrix Bernstein 
inequality~\citep[Theorem~1.4]{Tropp_2012_jour_FOCM},
which states that for independent, symmetric, random matrices
$Y_1,\ldots,Y_M\in\R^{n\times n}$ with $\E_H [Y_i] = 0$ and $\Vert Y_i\Vert_2\leq R$
almost surely, one has
\begin{align}
 \P\left( \left\Vert \sum_{i=1}^M Y_i \right\Vert_2 >t\right) \leq 
 n\exp\left(\frac{-{t^2}}{2\Vert\sum_i \E[Y_i^2]\Vert_2 +  \frac{2Rt}{3}}\right)
\label{eq_matrix_Bernstein}
\end{align}
for any $t>0$. Owing to the representation in~\eqref{eq_A_charac}, one can write
\begin{align*}
 \Dc^{-1/2}(A-\Ac)\Dc^{-1/2} = \sum_{e\in\Ec} (m-2)!
 \left(w_e-\E_H[w_e]\right)\Dc^{-1/2}R_e\Dc^{-1/2}
\end{align*}
as a sum of independent, zero mean random matrices.
One can verify that $\Vert R_e\Vert_2 \leq (m-1)$, and hence,
\begin{align*}
 \left\Vert (m-2)! \left(w_e-\E_H[w_e]\right)\Dc^{-1/2}R_e\Dc^{-1/2} \right\Vert_2 
 \leq \frac{(m-1)!}{\Dmin} \;.
\end{align*}
In addition, one can bound
\begin{align*}
&\left\Vert \sum_{e\in\Ec} \E_H\left[\left(
(m-2)! \left(w_e-\E_H[w_e]\right)\Dc^{-1/2}R_e\Dc^{-1/2}\right)^2 \right] \right\Vert_2
\\&= ((m-2)!)^2 \left\Vert \sum_{e\in\Ec} \Var_H(w_e)
\Dc^{-1/2}R_e\Dc^{-1}R_e\Dc^{-1/2} \right\Vert_2
\\&= ((m-2)!)^2 \left\Vert \sum_{e\in\Ec} \Var_H(w_e)
\left(\Dc^{-1}R_e\right)^2 \right\Vert_2
\\&\leq  ((m-2)!)^2 \max_{1\leq i\leq n} \sum_{j=1}^n \sum_{e\in\Ec} \Var_H(w_e)
\left(\left(\Dc^{-1}R_e\right)^2\right)_{ij}
\\&\leq  \frac{((m-2)!)^2}{\Dmin} \max_{1\leq i\leq n} \frac{1}{\Dc_{ii}} 
 \sum_{e\in\Ec} \E_H(w_e) \sum_{j=1}^n\left(R_e^2\right)_{ij} \;.
\end{align*}
Here, the first inequality holds due to Gerschgorin's theorem. Observing that the 
row sum of $R_e^2$ is at most $(m-1)^2$, the expression can be simplified to 
show that the quantity is bounded from above by $\frac{(m-1)!}{\Dmin}$.
Setting $t=3\sqrt{\frac{(m-1)!\ln n}{\Dmin}}$ in~\eqref{eq_matrix_Bernstein},
and combining above arguments, one arrives at~\eqref{eq_A_concentration}.
This completes the proof.

\subsection{Proof of Corollary~\ref{cor_ttm_sampling_examples}}
We begin by characterising $\beta$ for either sampling methods. Note that
$\beta \geq \max_e \frac{w_e}{p_e}$. Since, $|\Ec|=\binom{n}{m}$, it follows 
that for uniform sampling $p_e =  \binom{n}{m}^{-1}$ for all $e$, and hence,
an appropriate choice of $\beta = \binom{n}{m}$. On the other hand, for
the sampling in~\eqref{eq_ttm_wted_sampling}, $\max_e \frac{w_e}{p_e} = \sum_e w_e$.
Using Bernstein inequality, one may easily bound this term from above
by $2\sum_e \E_H[w_e] \leq 2\alpha_n\binom{n}{m}$, where the bound holds with probability
$(1-n^{-2})$. 

Thus, ignoring constants factors, one may set $\beta = \xi n^m$,
where $\xi=1$ for uniform sampling and $\alpha_n$ for weighted sampling.
The conditions in~\eqref{eq_dn_condns_sampled_example} follow directly
from~\eqref{eq_dn_condns_sampled} and $d,\delta$ computed in the proof of
Corollary~\ref{cor_ttm_consistency_balanced_dense}. 

\subsection{Proof of Lemma~\ref{lem_LLh_bound}}
Let $\beta$ be defined as in  Theorem~\ref{thm_ttm_consistency_sampled} and
$D_{\min} = \min\limits_{1\leq i\leq n} D_{ii}$. Assume that
\begin{equation}
 \Dmin > 36(m-1)!\ln n \quad\text{ and }\quad 
 N > 9 \left(1+\frac{2\beta(m-1)!}{\Dmin}\right) \ln n \;.
 \label{eq_dn_condns_sampled_restated}
\end{equation}
Also let $\Gamma$ denote the event
\begin{equation*}
 \Gamma = \left\{{D_{\min}}>\frac{\Dmin}2\right\}\bigcap\left\{\max_{e\in\Ec}\frac{w_e}{p_e} \leq \beta\right\}.
\end{equation*}
Then, conditioned on a given random hypergraph and the event $\Gamma$,
we claim that the following bounds hold with probability $(1-\frac{2}{n^2})$,
\begin{equation}
  \max_{1\leq i\leq n} \left| \frac{\Dh_{ii}}{D_{ii}} - 1\right|
 \leq 3\sqrt{\frac{\ln n}{N}\left(1+\frac{2\beta(m-1)!}{\Dmin}\right)} \;.
 \label{eq_D_concentration_sampled}
\end{equation}
and
\begin{equation}
\Vert D^{-1/2}(\Ah-A)D^{-1/2}\Vert_2
 \leq 3\sqrt{\frac{\ln n}{N}\left(1+\frac{2\beta(m-1)!}{\Dmin}\right)}\;.
 \label{eq_A_concentration_sampled}
\end{equation}
Assuming that the above hold,
we now derive a bound on $\Vert\Lh - L\Vert_2$ in the following way.
First, note that  the bounds in~\eqref{eq_D_concentration_sampled} and~\eqref{eq_A_concentration_sampled}
are with respect to a conditional probability measure, and need to be converted into a bound
with respect to the joint probability measure $\P_{S,H}$. This is not hard to derive
as one can see in the case of~\eqref{eq_A_concentration_sampled},
where one can write
\begin{align}
&\P_{S,H} \left(\Vert D^{-1/2}(\Ah-A)D^{-1/2}\Vert_2
 > 3\sqrt{\frac{\ln n}{N}\left(1+\frac{2\beta(m-1)!}{\Dmin}\right)}\right)
 \nonumber
 \\&=\E_H \left[ 
 \P_{S|H} \left(\Vert D^{-1/2}(\Ah-A)D^{-1/2}\Vert_2
 > 3\sqrt{\frac{\ln n}{N}\left(1+\frac{2\beta(m-1)!}{\Dmin}\right)}\right)\right]
 \nonumber
 \\&
 \leq\P_{H}\left(\Gamma\right)\E_{H|\Gamma} \left[ 
 \P_{S|H,\Gamma} \left(\Vert D^{-1/2}(\Ah-A)D^{-1/2}\Vert_2
 > 3\sqrt{\frac{\ln n}{N}\left(1\text{+}\frac{2\beta(m-1)!}{\Dmin}\right)}\right)\right]
 + \P_{H}\left(\Gamma^c\right)
 \nonumber
 \\&= O\left(\frac{1}{n^2}\right) + \P_{H}\left(\Gamma^c\right),
\end{align}
where the inequalities follow by observing that all the quantities are smaller than one,
and the first term is bounded due to~\eqref{eq_A_concentration_sampled}.
For bounding $\P_{H}\left(\Gamma^c\right)$, note that from~\eqref{eq_D_concentration}, it follows that
with probability $(1-O(n^{-2}))$, for all $i=1,\ldots,n$,
\begin{align*}
 D_{ii} > \Dc_{ii} \left(1 - 3\sqrt{\frac{(m-1)!\ln n}{\Dmin}}\right).
\end{align*}
Hence, if $\Dmin > 36(m-1)!\ln n$, then 
\begin{align*}
 D_{\min} > \Dmin \left(1 - 3\sqrt{\frac{(m-1)!\ln n}{\Dmin}}\right) > \frac{\Dmin}{2} \;.
\end{align*}
This fact, along with the assumption on $\beta$, shows that $\P_H(\Gamma^c)=o(1)$, 
and so, the upper bound on $\Vert D^{-1/2}(\Ah-A)D^{-1/2}\Vert_2$
holds with probability $(1-o(1))$ even with respect to joint probability measure. 
Similar result also holds for~\eqref{eq_D_concentration_sampled}.
Subsequently, we follow the arguments leading to~\eqref{eq_LLc_bound} to conclude that
\begin{align}
 \Vert \Lh - L \Vert_2 
 &\leq \max_{1\leq i \leq n} \left| {\frac{\Dh_{ii}}{D_{ii}}} - 1\right|
 \left(2+ \max_{1\leq i \leq n} \left| {\frac{\Dh_{ii}}{D_{ii}}} - 1\right|\right)
 + \Vert D^{-1/2}(\Ah-A)D^{-1/2} \Vert_2
 \nonumber
 \\&\leq 12\sqrt{\frac{\ln n}{N}\left(1+\frac{2\beta(m-1)!}{\Dmin}\right)} \;,
 \nonumber
\end{align}
where the last inequality holds with probability $(1-O(n^{-2}))$
under the conditions stated in~\eqref{eq_dn_condns_sampled_restated}.
To complete the proof, we derive the bounds~\eqref{eq_D_concentration_sampled} and~\eqref{eq_A_concentration_sampled},
which again rely on the use of Bernstein inequality.
For this, observe that
\begin{align*}
 \Dh_{ii} = \frac{(m-2)!}{N} \sum_{j=1}^n \sum_{e\in\Ic} \frac{w_e}{p_e} (R_e)_{ij}
 = \frac{(m-1)!}{N}\sum_{e\in\Ic} \frac{w_e}{p_e}\1\{i\in e\} \;,
\end{align*}
where for each $e\in\Ic$,
\begin{align*}
 \E_{S|H,\Gamma} \left[ \frac{w_e}{p_e}\1\{i\in e\} \right] &= 
 \sum_{e'\in\Ec: e'\ni i} p_{e'}\frac{w_{e'}}{p_{e'}} = \frac{D_{ii}}{(m-1)!} \;,
 \\ \Var_{S|H,\Gamma} \left[ \frac{w_e}{p_e}\1\{i\in e\} \right] &= 
 \sum_{e'\in\Ec: e'\ni i} \frac{w_{e'}^2}{p_{e'}} - \left(\frac{D_{ii}}{(m-1)!}\right)^2 
 \leq \left(\beta - \frac{D_{ii}}{(m-1)!}\right) \frac{D_{ii}}{(m-1)!}\;,
\end{align*}
and almost surely with respect to $\P_{S|H,\Gamma}$,
\begin{align*}
 \left| \frac{w_e}{p_e}\1\{i\in e\}-  \frac{D_{ii}}{(m-1)!}\right| 
 &\leq  \left(\beta + \frac{D_{ii}}{(m-1)!}\right) \;.
\end{align*}
Define $t = 3\sqrt{\frac{\ln n}{N}\left(1+\frac{2\beta(m-1)!}{\Dmin}\right)}$.
Since the samples $e\in\Ic$ are independent and identically distributed,
we can use Bernstein inequality to write
\begin{align*}
\P_{S|H,\Gamma}&\left( | \Dh_{ii} - D_{ii} | > tD_{ii} \right)
\\&= \P_{S|H,\Gamma}\left( \left| \sum_{e\in\Ic} \frac{w_e}{p_e}\1\{i\in e\} - \frac{D_{ii}}{(m-1)!} 
\right| > \frac{NtD_{ii}}{(m-1)!} \right)
\\&\leq 2\exp\left(\frac{-\frac{N^2t^2D_{ii}^2}{(m-1)!}}{
2N\left(\beta - \frac{D_{ii}}{(m-1)!}\right) \frac{D_{ii}}{(m-1)!} +
\frac{2}{3} \frac{NtD_{ii}}{(m-1)!}\left(\beta + \frac{D_{ii}}{(m-1)!}\right)}\right)
\\&\leq 2\exp\left(\frac{-\frac{ND_{ii}t^2}{(m-1)!}}{
\frac{2}{3}\left(4\beta - 2\frac{D_{ii}}{(m-1)!}\right)}\right)
\\&\leq 2\exp\left(\frac{-\frac{N\Dmin t^2}{2(m-1)!}}{
\frac{2}{3}\left(4\beta + 2\frac{\Dmin}{(m-1)!}\right)}\right) \leq \frac{2}{n^3} \;.
\end{align*}
The inequalities are derived using above relations, and the definition of $\Gamma$.
From above,~\eqref{eq_D_concentration_sampled} follows from union bound.

To prove~\eqref{eq_A_concentration_sampled}, observe from~\eqref{eq_Ah_defn} that
\begin{align*}
D^{-1/2}\Ah D^{-1/2} = \frac{1}{N}\sum_{e\in\Ic} (m-2)!\frac{w_e}{p_e} D^{-1/2}R_eD^{-1/2}
\end{align*}
is a sum of independent random matrices with
\begin{align*}
\E_{S|H,\Gamma} \left[(m-2)!\frac{w_e}{p_e} D^{-1/2}R_eD^{-1/2}\right] = D^{-1/2}AD^{-1/2} 
\end{align*}
and
\begin{align*}
\left\Vert(m-2)!\frac{w_e}{p_e} D^{-1/2}R_eD^{-1/2} - D^{-1/2}AD^{-1/2}\right\Vert_2
&\leq (m-2)!\beta \Vert D^{-1/2} R_e D^{-1/2} \Vert_2 +1 
\\&\leq \left(\frac{2\beta(m-1)!}{\Dmin}+1\right)\;.
\end{align*}
The first bound uses the fact $\Vert D^{-1/2}AD^{-1/2}\Vert_2 =1$ and the second
follows since $D_{\min} > \frac12\Dmin$ and $\Vert R_e \Vert_2 \leq (m-1)$.
We can also bound the norm of the variance term as
\begin{align*}
&\left\Vert \E_{S|H,\Gamma} \left[\left((m-2)!\frac{w_e}{p_e} D^{-1/2}R_eD^{-1/2} 
- D^{-1/2}AD^{-1/2}\right)^2\right]\right\Vert_2 
\\&=\left\Vert - \left(D^{-1/2}AD^{-1/2}\right)^2 +
((m-2)!)^2\sum_{e\in\Ec}\frac{w_e^2}{p_e} D^{-1/2}R_eD^{-1}R_eD^{-1/2} \right\Vert_2
\\&\leq 1 + \frac{((m-2)!)^2\beta}{D_{\min}}\left\Vert \sum_{e\in\Ec} w_e  D^{-1}(R_e)^2\right\Vert_2
\leq \left( 1 + \frac{2\beta(m-1)!}{\Dmin}\right)\;.
\end{align*}
Using these relations and the matrix Bernstein inequality, the bound in~\eqref{eq_A_concentration_sampled} 
can be derived quite similar to the derivation of~\eqref{eq_A_concentration}. 

\vskip 0.2in
\bibliography{graph_learning,hypergraph_learning,planted_model,tensor_theory,other_theory,experiment,new_ref}

\end{document}